\documentclass[letterpaper,12pt]{report}
\usepackage{thesis}
\usepackage{graphicx}
\usepackage{tabularx}
\usepackage{booktabs}
\usepackage{svg}
\usepackage{algorithm}
\usepackage[noend]{algpseudocode}
\usepackage{amsmath}
\usepackage{amssymb}
\usepackage{soul}

\begin{document}
\pagenumbering{roman}


\thesistitle
        {Improving Robotic Manipulation: Techniques for Object Pose Estimation, Accommodating Positional Uncertainty, and Disassembly Tasks from Examples}
	{\emph{Viral Rasik Galaiya}}
	{Master of \emph{ Science}}
        {Supervisor: Dr. Vinicius Prado da Fonseca\\
        Dr. Xianta Jiang}
	{Department of \emph{Computer Science}}
	{\emph{July 2024}}

\addcontentsline{toc}{chapter}{Abstract}
\begin{center}
\textbf{\large Abstract}
\end{center}

To use robots in more unstructured environments, we have to accommodate for more complexities. Robotic systems need more awareness of the environment to adapt to uncertainty and variability. Although cameras have been predominantly used in robotic tasks, the limitations that come with
them, such as occlusion, visibility and breadth of information, have diverted some focus to tactile sensing. In this thesis, we explore the use of tactile sensing to determine the pose of the object using the temporal features. We then use reinforcement learning with tactile collisions to reduce the number of attempts required to grasp an object resulting from positional uncertainty from camera estimates. Finally, we use information provided by these tactile sensors to a reinforcement learning agent to determine the trajectory to take to remove an object from a restricted passage while reducing training time by pertaining from human examples. 
\addcontentsline{toc}{chapter}{Acknowledgements}
\begin{center}
\textbf{\large Acknowledgements}
\end{center}

I would never have even started my master's program if it were not for Vinicius' enthusiasm to share the work he was doing after a chance encounter with him while rock climbing. I'd like to thank my supervisors, Dr. Vinicius Prado de Fonseca and Dr. Xianta Jiang, for their invaluable support, guidance, and patience. 

I would also like to acknowledge our collaborator, Dr. Thiago Eustaquio Alves de Oliveira, for his ideas and advice on the work I have done, as well as Mohammed Asfour for his input in the second chapter. 

This would not have been possible without the encouragement of my family, Rasik Galaiya, Ila Galaiya and Kangana Galaiya. 
Many thanks to Yanitza Trosel for all her support and advice and her mighty sidekick Suki.

A shoutout to my friends in Computer Science, Marius Seidl, for the engaging conversations, and Elahe Mohammadreza for the camaraderie as well as all my other lab mates and colleagues. 

Finally, all the friends I made in Newfoundland who helped make it my new home; Neus Campanya Llovet, Celyn Khoo, John Green, Haris Mubashir, Pranjal Patra, and Ali Ghamartale.

\tableofcontents

\addcontentsline{toc}{chapter}{List of Tables}
\listoftables

\addcontentsline{toc}{chapter}{List of Figures}
\listoffigures

\pagenumbering{arabic}
\chapter{Introduction}
\label{1_overview}

\section{Background}

Outsourcing human labor to a being or device has been conceptualized since as early as the Greek-Hellenistic age \cite{Gasparetto2019}, while there are even records of a musical automaton in the 1100s \cite{iavazzo2014evolution}. Although the idea of humanoid robots has been around in various ages and cultures ranging from Roman mythology to ancient India, the widespread robotics application in the 1950s was machine tools combined with manipulators to perform simple, repeatable activities without information about the environment. With time, these manipulators started becoming more sophisticated, incorporating advances in machine learning, sensor technology \cite{zamalloa2017}, and mechanical and material advancements to come full circle to humanoids. These first implementations were very rudimentary compared to today's standards because of the lack of information on the environment and were operated in highly structured environments. With the development of sensors, computing, and application breadth, robotic applications keep moving closer to unstructured environments. The ability to accurately infer the environment and adapt to this variable environment plays a significant role in robots' wider range of uses. Zamalloa \textit{et al.} \cite{zamalloa2017} describes the trend as moving from mass manufacturing to mass customization.

 Since the development of robotics, the breadth and depth have increased tremendously. Despite manipulators being the first at mass adoption \cite{zamalloa2017} and becoming just a subset of the whole field that now includes drones and quadrupeds, it is still a highly researched area due to the range challenges still far from being solved like gripper design \cite{kumar2023design}, control \cite{Liu2021} and sensing \cite{dou2021soft}.

Humans have an extensive range of sensing capabilities that enable us to perform very complex tasks seemingly efficiently. We rely highly on visual perception systems but also have other very powerful sensory and information input systems. These systems include sight, touch, sound, and smell. These systems can be used very well together but are also used well in isolation. Helbig \textit{et al.} \cite{helbig2008haptic} suggests that since no information-processing system has the ability to provide sufficient information under all conditions, it is imperative to combine multiple systems. This line of thought has encouraged researchers to explore forms of perception systems other than vision directly, ranging from inertial measurement units \cite{Li2020} to rangefinders \cite{Alatise2020}. In particular, vision has limited capabilities for several grasping and manipulation tasks, specifically for near contact and contact activities. This is because we can't generally obtain information on weight, force, and friction with an unmodified vision system. Albeit, creative methods of using cameras have been invented to obtain tactile information \cite{wang2021gelsight}. 

Despite the wide range of sensor data as well as the quantity of data, there is intrinsic uncertainty from the sensor's accuracy, geometric uncertainty from sensor limitations like placement or types, physical uncertainty from properties such as material composition or weight as well as environmental disturbances from the environment, especially in unstructured environments. Reinforcement learning with tactile sensing allows for adapting to these uncertainties and changing environments \cite{dong2021tactile}.  

\section{Literature Review}

For any manipulation task, it is vital to know the pose of the object being manipulated. It is also important to keep track of changes that are made when manipulating the object so the trajectory of the manipulator can be efficiently calculated. Amongst various other fields, object pose estimation has benefited highly from the advent of deep learning techniques. Du \textit{et al.} \cite{DuGuoguang2021Vrgf} explores work done in 2D planner grasps six-degree grasp planning including constraint grasps, object pose estimation using template matching-based methods, voting-based methods, end-to-end grasp estimation using both RGB-D images. They concluded that there needs to be a close similarity of the 3D object model the network is trained on to have accurate grasps. Moreover, the lack of geometry information from the direction not facing the camera affects the accuracy of the shape completion and suggests the use of multi-source data. Therefore, other data sources have been explored, especially close to or in relation to the contact points, such as using tactile sensing or proprioception \cite{Sipos2022}. These methods generally complement vision-based sensing modalities to address blind spots, increase accuracy or both. Various ways of leveraging the information of tactile sensing for pose estimation have been used, including collisions \cite{Li2021, contact_pose}, simulation to real life \cite{bauza2023tac2pose, Yang2024} and reinforcement learning \cite{kelestemur2022tactile}. 

Grasping in unstructured environments must overcome significant uncertainties from things like occlusion and noise and accommodate many object properties like geometry and mass. Under-actuation and compliance are great low-cost rectifiers. This is because fewer motors are required to grasp an object, and compliance increases adaptability, reducing the effects of uncertainty. Although these techniques can result in position error because of forces dislodging the object, they can be reconciled using information obtained from the end-effector, such as tactile sensors \cite{Jentoft2014}. Various compliant tactile sensors \cite{Xiong2022, Fernandez2021, Liu2022, Oliveira2017, cretu2015computational, de2023bioin} have been developed to improve grasp force control, prevent slip, improve pose estimation and accommodate positional uncertainty. Kuppuswamy \textit{et al.} \cite{Kuppuswamy2020} use a combined visio-tactile approach to update visual estimates of the object pose such that it converges with less than 15 touches. In contrast, Gasparetto \textit{et al.} \cite{Gasparetto2019} uses tactile information for global localization of an object using a convergence based on local geometry. Jiang and team \cite{Jiang2023} used a visio-tactile based approach as well; the camera providing a region to poke transparent objects and obtain the local information that can be used to increase the likelihood of a successful grasp. 

Once the object has been grasped, the object can be used for assembly tasks such as peg-in-hole, slide-in-groove, bolt screwing, pick and place, and pipe connection \cite{jiang2020state}. Although peg-in-hole \cite{dong2019tactile,dong2021tactile, Shiyu2021, SONG20211, Lee2022} is a commonly explored topic, most of the research focuses on collision with the edges and aligning the object with the hole. Dong \textit{et al.} \cite{dong2021tactile} compared the performance of insertion policies using tactile RGB and tactile flow. They also compare the performance of a tactile sensor and a torque/force sensor with a vision-based tactile sensor. Finally they compare supervised learning and reinforcement learning. Testing with 4 different objects, using the tactile sensor with a tactile flow presentation and reinforcement learning provided the best results. Various methods have been investigated for insertion, such as using perturbations \cite{Lee2022}, uncertainty distribution exploration \cite{Kang2022}, reinforcement learning, and supervised learning \cite{kim2022}.

\section{Motivation}

In-hand object position is important knowledge for various complicated tasks such as repositioning the object, collaborating with other robots or humans and interacting with the environment. However, these tasks result in changes in the object position with these interactions, and it can be beneficial to keep track of these changes or identify the changes after the interaction to adapt to them and perform tasks based on accurate and updated information. Moreover, when these interactions occur, the proximity to the object is small, the likelihood of the source of the interaction is high, and the obscuring cameras are more likely to occur in these situations. Hence tactile sensors may be used to predict and update the position. Moreover, since these motions are sequential, taking into account the temporal aspect of the signals from the tactile sensors can assist in increasing accuracy.  

This occlusion can also occur by the manipulator during approach when grasping. The accuracy of the object pose obtained from the camera can be affected by varying light conditions and object characteristics like colour and environmental characteristics like the wind. The use of tactile sensors and adaptive algorithms can compensate for these factors by updating the internal state of the error from the interactions, which are both successful and unsuccessful.

These problems associated with occlusion are also prevalent in disassembly tasks because parts of the objects that need to be disassembled are generally occluded, and those parts influence the trajectory required, such as the peg-in-hole. These limitations can be accommodated by using adaptive algorithms like reinforcement learning. In particular, the disassembly trajectory is updated based on the interactions with the environment. Moreover, the system will take a shorter time to improve its performance if it is being implemented over a previously trained system, for instance, from human demonstrations.





\section{Objectives}

Based on this motivation, we defined the following objectives:

\begin{itemize}
    \item Use tactile sensors to estimate the object pose after grasp. Study deep learning to take advantage of the temporal information obtained from tactile sensors to predict object pose based on external forces.
    \item Develop methods to estimate object pose by using reinforcement learning to adapt to uncertainties using tactile feedback.
    \item Evaluate the effect of pre-training on reinforcement learning in extracting objects from a peg-in-hole setup. 
\end{itemize}

The goal of this thesis is to explore the usage of tactile sensing in the various components of robotic manipulation. Specifically, to improve grasping, obtain pose estimations of objects in hand and guide the extraction process of the peg-in-hole problem.

\section{Contributions}
pretraining In this thesis, we use a long short-term memory network with tactile information to predict the object's angle. We subsequently use reinforcement learning and a compliant tactile sensor to address pose uncertainty from camera inputs. Finally, we use a pertaining with human examples strategy with reinforcement learning to speed up the learning phase of the agent to extract the peg in the peg-in-hole setup. 

The work of this thesis has been published partially in other versions in peer-reviewed conferences and journals. We have the approval of the publishers to use the published findings in the thesis. Published versions of this research Chapter:

\begin{itemize}
    \item Chapter \ref{chap:ch3_pose_estimate} explores the use of temporal tactile data via a sliding window to predict the object angle under grasp. This paper is published as \textbf{Galaiya VR, Asfour M, Alves de Oliveira TE, Jiang  X, Prado da Fonseca V. Exploring Tactile Temporal Features for Object Pose Estimation during Robotic Manipulation. Sensors. 2023; 23(9):4535.} \cite{Galaiya2023}
    \item In Chapter \ref{chap:ch4_pick_objects}, we devise a policy that effectively models object position estimation errors and reduces exploratory sensor contact, improving the grasp execution process. This paper is approved as \textbf{Galaiya VR, Alves de Oliveira TE, Jiang  X, Prado da Fonseca V. Grasp Approach Under Positional Uncertainty Using Compliant Tactile Sensing Modules and Reinforcement Learning. Sensors. 2024 (accepted)}

\end{itemize}

\section{Thesis Outline}

In Chapter \ref{chap:ch3_pose_estimate}, we explore the effects of incorporating the temporal data of tactile information for pose estimation. We use a long short-term memory network to predict the object's angle and compare it to standard regression models that do not utilize the time series information. Chapter \ref{chap:ch4_pick_objects} explores the use of reinforcement learning and a compliant tactile sensor to address pose uncertainty from camera inputs. Finally, in Chapter \ref{chap:ch5_pull_objects}, we explore the use of tactile sensors in disassembly tasks using reinforcement learning.

\chapter{Exploring Tactile Temporal Features for Object Pose Estimation
During Robotic Manipulation}
\label{chap:ch3_pose_estimate}
\section{Introduction \label{sec:intro}}
Many areas of human activity, such as mass-production factories, minimally invasive surgeries, and prostheses have adopted robotic manipulation systems.
Robotic manipulation is exceptionally reliable when the system has complete information regarding the environment.
These systems usually must follow a set of trajectories, interact with objects of known features, and perform repetitive tasks with minimal environmental adaptation, which limits the use of manipulation systems to perform activities in unstructured settings.
Recent advancements in data-driven methods, innovative gripper design, and sensor implementation have reduced the limitations of robotic manipulators in such environments.
Nevertheless, there are hurdles to the applications of robotic arms in unstructured environments or dexterous tasks such as complex manipulation of daily objects \cite{billard2019}.

One main challenge is estimating the object's orientation during the after-grasp phase.
The object's orientation can change from an initial visual estimation due to calculation errors, external forces, finger occlusion, and clutter.
After a successful grasp, one approach is to use tactile sensors to extract object information, improving the object's pose estimation.

Robotic hands have immense flexibility despite their use in specific domains, such as prostheses, with limitations regarding the human hand's size, weight, and shape.
By sacrificing initial stability and uncertainty in grasp pose estimation, an under-actuated approach substantially reduces planning time and gripper design complexity \cite{Hammond2012}. 
However, it is fundamental for robotic arms to estimate the handled object's pose to operate optimally in object manipulation applications.
For instance, the grasp used by a gripper of a robotic arm or a prosthesis to hold a mug might change if its handle is at a different angle.

Object orientation estimation depends on several aspects, such as the gripper's configuration, the sensors used, and how the data is analyzed. 
Different sensors are used in robotic manipulation to categorize the properties of an object, such as its orientation.
Pre-grasp poses are commonly obtained using computer vision \cite{SAHIN2020103898}.
However, having visual data only can be insufficient as the gripper approaches the object and the range of occlusion increases.
This limitation is particularly pronounced when the camera's location is fixed or under unpredictable circumstances, i.e., in unstructured environments.
For instance, using a top-view camera to estimate the object pose is not feasible for an arm prosthesis, whereas prosthesis-mounted cameras are susceptible to occlusion.
Moreover, once the gripper grasps the object, it will cover at least part of it, making it difficult to estimate its orientation.
Furthermore, merely using vision cannot reduce forces and related environmental stimuli, leading to potential errors in the estimation of orientation due to miscalculated geometry, friction, forces, camera occlusion, and clutter \cite{billard2019, Kroemer2019}. 

Due to the limitations of visual methods, several applications use tactile sensing \cite{wettels2008biomimetic, ward2018tactip, lambeta2020digit, alspach2019soft, su2015force, yoon2022Elong} while grasping the object, providing more relevant information that is not interrupted \cite{Li2020}. Tactile sensing has shown promise in specific use cases, such as in minimally invasive surgery {\cite{Bandari2020}} or cable manipulation {\cite{she2021cable}}, and is also being shown to be a good supplement to control system optimization {\cite{bi2021zero, zhang2023interaction}}.
Sensors such as pressure sensors {\cite{yoon2022Elong}}, force sensors {\cite{VonDrigalski2020}}, and inertial sensors {\cite{Oliveira2017}} are gradually becoming more prevalent for object pose estimation and object recognition. In addition, tactile sensors provide after-grasp contact information about the object that can be used for control {\cite{she2021cable}} or in-hand manipulation. Nevertheless, there have also been developments of vision-based tactile sensors ranging from using internal reflection {\cite{Gomes2020, Romero2020}} to observing the deformation of the surface {\cite{Trueeb2020}}.

Tactile sensing can be a vital link to overcoming computer vision limitations and can result in better performance of robotic manipulation.
Previous works have used machine learning models and visual frames of reference to train models that learned the after-grasp object angle, which is later used to estimate the object's pose \cite{PradodaFonseca2019}.
However, previously seen data can affect the current estimation of the object's state.
So, we hypothesize that estimating the current object pose can be improved by considering temporal data, such as when sliding window sampling.
For this reason, in the present work, we study the effect of temporal data based on sliding window sampling to train a deep learning model for object angle estimation.

\section{Literature Review}
Orientation estimation has been a part of pose estimation in robotics research for a long time. Recent studies have made leaps regarding orientation estimation with sufficiently low error due to advancements in sensors technology, most importantly tactile sensors {\cite{Sipos2022sim, Lloyd2022goal, Lima2019zscore, de2015data}}.

For instance, Ji \textit{et al.} {\cite{new_ji_tactile}} proposed a novel model-based scheme using a visual-tactile sensor (VTS) {\cite{new_vts}}. In their study, the sensor compromised a deformable layer that interacted with objects with a depth camera behind said layer to generate a depth map of the deformation caused by the object. They reported orientation errors for three objects under $3 ^{\circ}$. However, detecting their objects' rotations could have been visually easier compared to more uniform smooth shapes such as cylinders or ellipsoids.

Additionally, Suresh \textit{et al.} {\cite{new_suresh_slam}} formulated the tactile sensing problem as a simultaneous localization and mapping (SLAM) problem, in which the robot end effector made multiple contacts with the object to determine its pose. They reported a rotational root mean square error (RMSE) of 0.09 radians. However, their method assumed the initial pose and scale of the object roughly and neglected factors outside the controlled setup that might change the object's orientation.
\newline\newline
Other studies utilized information about the robot arm alongside tactile data to estimate the orientation of objects. Alvarez \textit{et al.} \cite{alvarez} used the kinematic information and a particle filter for pose estimation via tactile contact points, force measurements, and angle information of the gripper's joints.
Their algorithm initiates a pose estimation using visual data, which is refined by a particle filter based on the optical data. After experiments with three objects of different sizes, they report a $0.812^{\circ}$ error in their best experiment, which rises to $3.508^{\circ}$ in their worst case. Results aside, the method requires a known kinematic model of the robotic arm and a top-view camera for inference, which is infeasible in some applications, like daily activities using prostheses.
\newline\newline
To relax the requirement of a detailed kinematic model, recent research has explored underactuated grippers while relying on machine learning methods to build a model of the object pose.
For instance,  Azulay \textit{et al.} \cite{Azulay} conducted a wide-scope study to investigate objects' pose estimation and control with under-actuated grippers.
They incorporated haptic sensors, joint angles, actuator torques, and a glance at the pose at the start of the gripper's movement.
Using the robotic arm's kinematic model, they concluded that some combinations of the tested features are better suited for object manipulation than others.
They report a root mean square error (RMSE) of $3.0\pm 0.6^{\circ}$ for orientation using a neural network with LSTM layers, their best model. Using multiple features alongside the kinematic model can be computationally intensive for processors on devices such as prosthetics.

However, investigating orientation estimation itself only prior to grasping can limit the reported results in some situations. For instance, robotic grippers that handle objects can occlude the object partially or fully, affecting visual-based approaches. Furthermore, objects can rotate during handling due to many factors, such as slipping or external forces, thus requiring methods to estimate objects' orientation during the grasp phase.

High-density tactile sensors, akin to the human hand, are another direction that can provide much information. Funaabashi \textit{et al.} {\cite{Funaabashi2022multi}}  used graph convolution neural networks (GCNs) to extract geodesical features from 3-axis tactile sensors across 16 degrees of freedom of a robotic hand providing 1168 measurements at 100Hz. They used eight objects with two different hardness, slipperiness, and heaviness factors. They compared various GCN configurations and a  multilayer perceptron, and the GCN model with the most convolution layers was the best performer. The limitation of this method is due to the need for high computational resources, the requirement of a large number of sensors, and the ambiguity of intermediate states, although accounting for different properties, such as hardness and slipperiness, improves possibilities of generalization. 
\newline\newline
To develop a solution that required minimal finger path planning, relaxed kinematic model requirement, and less needed processing of images, Da Fonseca et al. {\cite{PradodaFonseca2019}} developed an underactuated gripper with four compliant sensing modules on flexible fingers, and investigated the collected sensor data while grasping objects of three distinct sizes. 
The experiments included a top-view camera to obtain a visual frame of reference for ground truth orientation.
The method used tactile sensors' information to represent the object angle, whereas the ground truth angle was obtained from the camera frame.
Finally, the authors compared five regression models trained using tactile data to estimate the object's angle.
The best models reported by the authors were the ridge regression model and linear regression, obtaining a $1.82^{\circ}$ average mean square error.
The authors used random data sampling for model training in the paper and left possible relationships among the time-series samples as a future research point.
Still, given that the tasks were dynamic, we expect that the near samples in the time-series sensor data are correlated with the angle.
\newline\newline
Some studies also investigated the fusion of tactile and visual data for orientation estimation during object handling.

Alvarez \textit{et al.} {\cite{new_alvarez_fusion}} proposed a fusion method of the visual data and tactile data to estimate the object's pose during grasp. A camera tracked the object during grasp, whereas a particle filter was utilized with the tactile data to reduce the uncertainty of the object's pose. They reported that their method obtained an orientation error varying from $1^{\circ}$ to $9.65^{\circ}$. Their method yielded a high variance of the estimation error, in addition to requiring a 3d model of the handled object for the method to be used.

Dikhale \textit{et al.} {\cite{new_dikhale_inhand}} proposed sensor fusion of visual and tactile data as well. Their method used neural networks to process the tactile and visual data separately before fusing them to give a final prediction of the object's pose. They reported an angular error as small as $3^{\circ}$; however, it reached a high of $24^{\circ}$, showing high variance in the estimation depending on the object.
\newline\newline

From the previous studies, we find that different factors affect the orientation estimation performance and eligibility. For instance, computationally-demanding methods, such as ones relying on inverse kinematics or particle filters, are inappropriate for small devices, such as prostheses limited to an onboard processor. Whereas relying on visual data, solely or with sensor fusion, is prone to occlusion during the grasp phase as top-view cameras are not feasible in many applications. Hence, a model must only estimate the object's angle using only tactile data during grasp without kinematics to reduce computation while providing an acceptable angle error.

We evaluate the use of sliding window sampled tactile data to estimate the yaw angle under the stable grasp of an object while relaxing the kinematic model requirement by using an underactuated gripper, and a compliant bio-inspired sensing module that includes magnetic, angular rate, gravity, and pressure sensing components. 
We analyze the temporal nature of tactile signals by using a neural network that contains long short-term memory (LSTM) layers to estimate the orientation with the highest precision for objects.
The models trained in the present work were based on Da Fonseca \textit{et al.} \cite{PradodaFonseca2019}, taking in a window of readings from the sensors mounted on the gripper and then outputting the estimated object's orientation at the end of this sampling window.
As the chapters main topic is the in-hand orientation estimation, our method uses only the initial grasp orientation as a reference and does not require information from the gripper joints, its kinematics model, a multitude of sensors, nor during-grasp visual data. 

Our method can be utilized in a multitude of applications from everyday use to factory settings due to its dependency on only a small number of tactile sensors without the need for additional types of sensors. Furthermore, our method does not need computationally-capable machines as it utilizes only a neural network that can run on a computational device as small as a flash drive, such as Google Coral, due to advancements in computational technology. In addition, the proposed method's performance is limited to uniform shapes whose orientation change is hard to determine visually, such as rotating cylinders.

\section{Materials and Methods}

Here we describe the data collection and pre-processing methods used for sliding window-sampling tactile data, the models trained for the experiments, and how we organized the sampling strategy for pose estimation.

\subsection{Data Collection\label{sec:data-colect}}
We used tactile data collected in a previous study \cite{PradodaFonseca2019} from an underactuated gripper with two independently controlled fingers during object-grasping tasks to evaluate the sliding window sampling strategy for pose estimation.

\begin{figure}[!ht]
    \centering
    \includegraphics[width=0.5\textwidth]{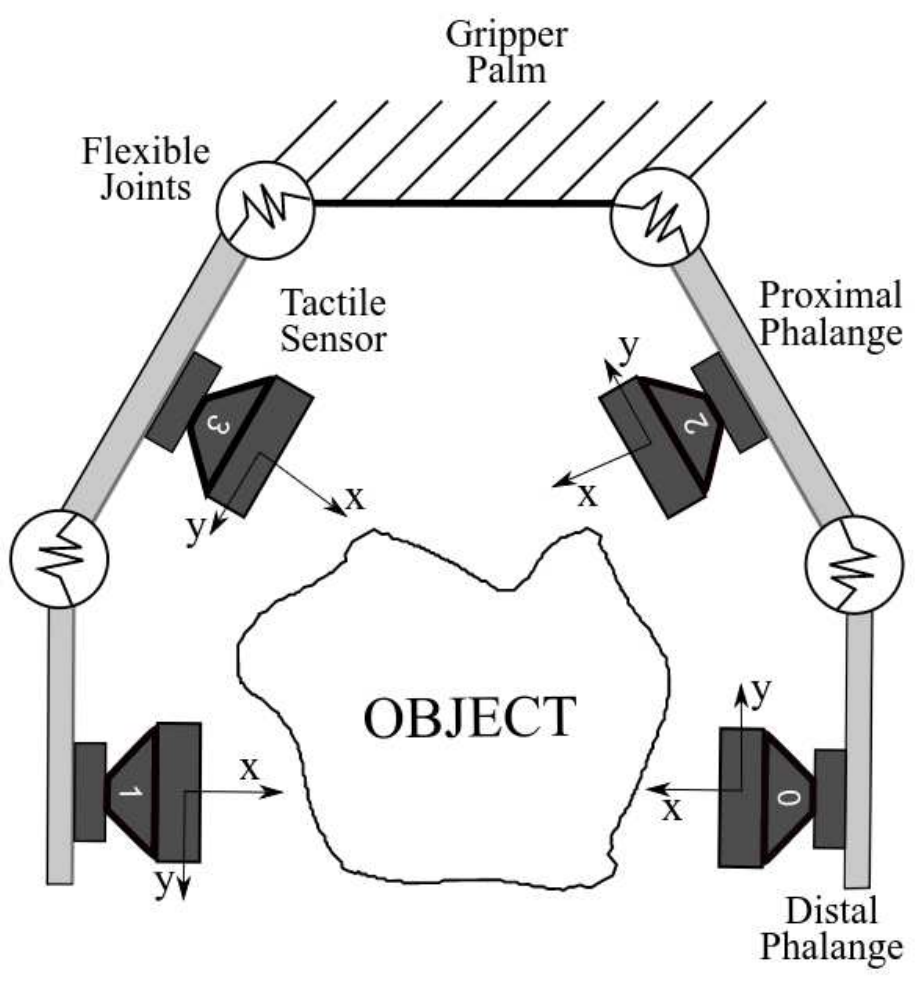}
    \caption{The underactuated gripper \cite{PradodaFonseca2019} diagram with two fingers, each with two phalanges and their respective sensors.}
    \label{fig:GripperModel}
\end{figure}

In the gripper developed by Prado da Fonseca \textit{et al.} ~\cite{PradodaFonseca2019}, each phalanx has a fixed tactile sensor developed by Alves de Oliveira \textit{et al.} ~\cite{Oliveira2017}, as shown in Figure~\ref{fig:GripperModel}.

\begin{figure}[!ht]

    \begin{minipage}[b]{\columnwidth}
        \centering
    \includegraphics[width=0.6\columnwidth]{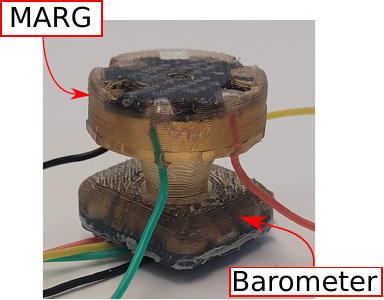}
    \caption{The sensor with its base attached to the manipulator and close to where the barometer is, and its surface over the MARG sensor is in contact with the object.}
    \label{fig:marg_sensor}
    \end{minipage}
\end{figure}

Each sensor provides deep pressure information from a barometer in addition to angular velocity, linear acceleration, and magnetic field in all three axes using the 9-degree-of-freedom magnetic, angular rate, and gravity (MARG) system. The barometer as shown in figure {\ref{fig:marg_sensor}} is encased in a polyurethane structure close to the base, and the MARG sensor is placed closer to the point of contact so it can detect micro-vibrations. The fabrication structure of the sensor enables the pressure to be transferred from the contact point to the barometer effectively.
The compliant sensor structure allows the contact displacement to be measured by the inertial unit while the deep pressure sensor measures the contact forces.
The data is collected using an onboard micro-controller interfacing via I2C with a computer running the Robot Operating System (ROS) framework \cite{ros}. 

\begin{figure}
    \centering
    \includegraphics[width=0.5\textwidth]{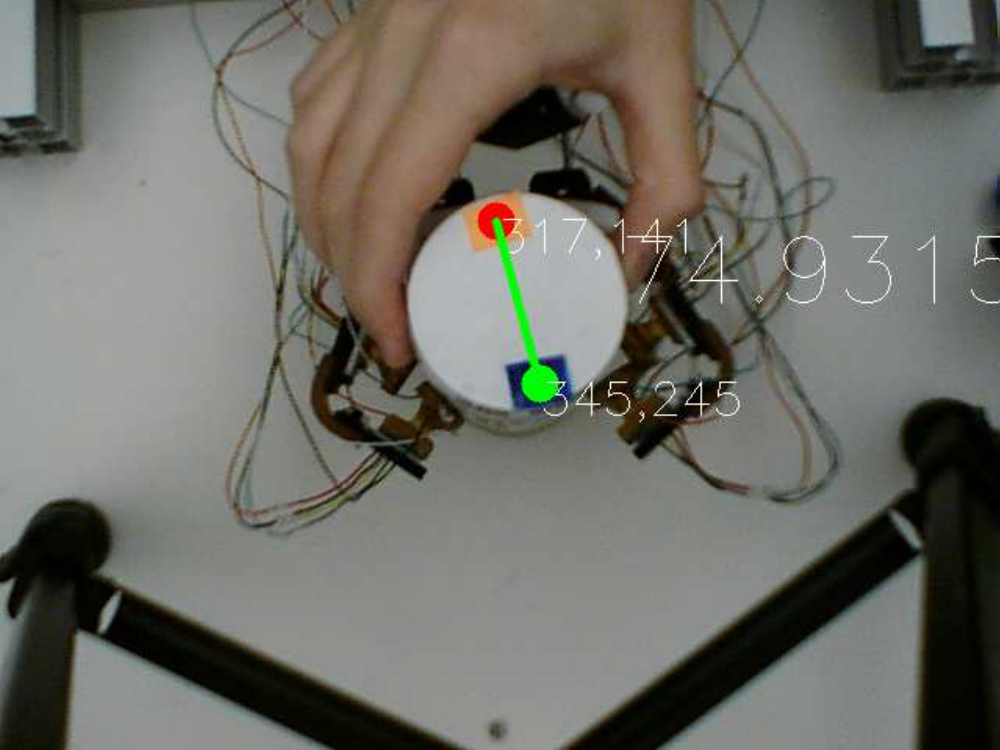}
    \caption{The object's two markers to obtain the ground truth angle using computer vision \cite{PradodaFonseca2019}.}
    \label{fig:gripper-setup}
\end{figure}

Prado da Fonseca \textit{et al.} \cite{PradodaFonseca2019} used the allocentric reference frame from the camera pointed down to calculate the object's angle. The top view angle of the object is extracted using two colored markers attached to it to identify key points using the OpenCV library as shown in Figure~\ref{fig:gripper-setup}.

The angle between this the two markers line and the fixed camera frame horizontally in the clockwise direction is established to be the object's angle, and the object is considered at $90^\circ$ on the $x-axis$.
These points are later compared to a fixed frame of reference at the camera's center to determine the object position change relative to the specified frame of the gripper.
The stable grasp was obtained using a dual fuzzy controller that obtained micro-vibrations and pressure feedback from the tactile sensor {\cite{PradodaFonseca2022tactile}}.
 This procedure was performed with three cylindrical objects with $57$ mm, $65$ mm, and $80$ mm diameters. The objects were rotated manually in CW and CCW directions, simulating external forces causing the object to change its orientation during grasp. 
 Although this motion is at a low speed, the human element of this motion provides inconsistent forces, which the model was able to take into account to provide an accurate prediction.
Such movements also simulate the act of parasitic motions, which are undesired motion components which lead to lower manipulation accuracy/quality {\cite{nigatu2021analysis}}, 
despite being in stable grasp. Moreover, the three different objects are used to determine the ability of the model to generalize among similar objects. The ground truth angle after rotation is obtained relative to the form of reference from the top view camera as seen in Figure {\ref{fig:gripper-setup}}.

\subsection{Data Characteristics} 

The pre-processing methods used in this work depend highly on the time-series details of the data available from Prado da Fonseca \textit{et al.} \cite{PradodaFonseca2019}. 
For instance, the number of instances in each window sample can be affected by the different frequencies of each sensor. 
Table~\ref{tab:table-sensor-frequency} shows the average sampling frequency of each sensor, in which the slowest sensor is the camera, ranging from $9$ to $29.95\ Hz$. The fastest sensor is the MARG sensor, ranging from $911.33$ to $973.50\ Hz$. 

\begin{table}[!ht]
    \centering
    \caption{The average frequency of the data obtained from its respective sensors.}
        \begin{tabularx}{\textwidth}{ccc}
    \toprule
    \textbf{Camera} & \textbf{Pressure} & \textbf{MARG sensor} \\
    \midrule
    29.95Hz & 402.19Hz & 973.50Hz \\  
    \bottomrule
    \end{tabularx}
    \label{tab:table-sensor-frequency}
\end{table}

\begin{figure}[!ht]
\centering
\includegraphics[width=1\textwidth]{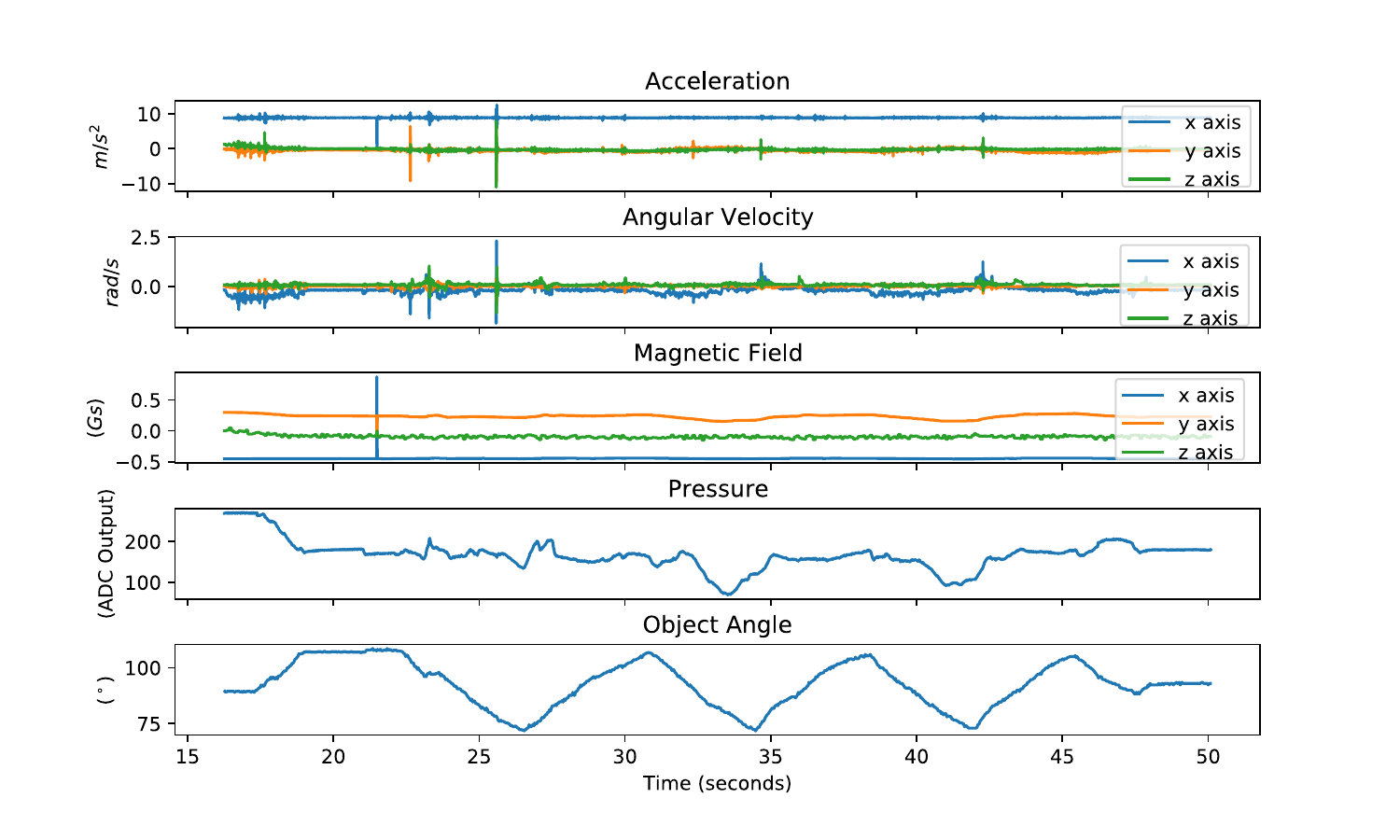}
\caption{The angle rotation, the corresponding pressure, and one of the four MARG outputs in one of the trial data collection trials.}
\label{fig:SensorReading}
\end{figure}

As mentioned in Section~\ref{sec:data-colect}, the data collection consisted of a CW and CCW rotation procedure performed by an external operator on three cylindrical objects with $57$ mm, $65$ mm, and $80$ mm diameters. 
The dataset for each object contains sensor readings from five different external rotation operations.
Figure~\ref{fig:SensorReading} shows the disturbances of rotation on the pressure, linear acceleration, angular velocity, and magnetic field for one sensor in relation to the angle during external rotations.

The data characteristics described here are sufficient for our investigation. Further details about the data collection protocol and attributes can be found in the original data collection study
\cite{PradodaFonseca2019}.

\subsection{Pre-processessing}
\begin{figure} [thb!]
    \centering
    \includegraphics[width=\textwidth]{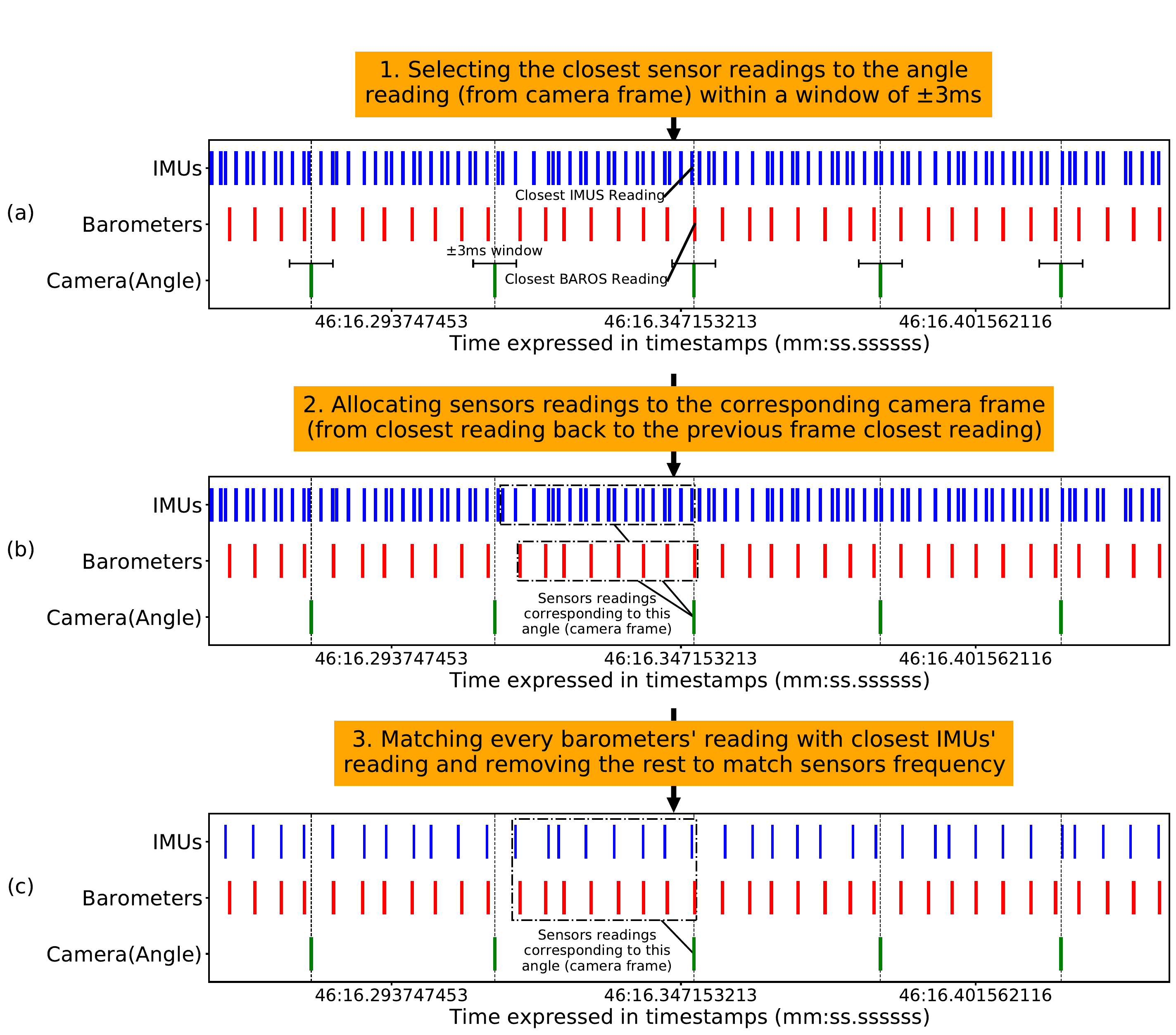}
    \caption{The procedure used for sensors alignment. (a) For every angle determined by the camera, the closest corresponding pressure and MARG values were selected. (b) Sensor values were grouped with the corresponding camera frame. (c) Downsampling MARG values to match the frequency of the pressure sensor.}
    \label{fig:allgiment}
\end{figure}

\begin{algorithm}
\caption{Pre-processing and experimentation pseudo code}
\begin{algorithmic}[1]
\For{each barometer reading
}:
    \State keep closest MARG reading
\EndFor
\State discard the rest of the MARG reading

\For{For each angle value}:
    \State take sensor readings of corresponding timestamp
    \State take $(WindowSize -1)$ previous sensor readings
\EndFor

\State separate training and test data
\State normalize training and test sensor values using the mean and standard deviation from training data

\State train model using training data
\State obtain performance results using test data
\end{algorithmic}
\end{algorithm}
The listener ROS node collected data at different time instances since the camera, MARG sensor, and pressure communicated asynchronously.
Therefore the signals needed to be aligned for our strategy of window sampling. First, we scaled the data to utilize deep learning methods.
Subsequently, to add LSTM layers, we had to reconcile the sampling frequency differences for the various sensors by synchronizing and downsampling their data. Since the lowest frequency was the camera frames, their timestamps acted as a reference for our procedure of sensor alignment presented in Figure~\ref{fig:allgiment}. Afterward, we reshaped the data to incorporate the previous states for each instance of the ground truth angle.

Figure~\ref{fig:allgiment} (a) shows that the obtained pressure and MARG signals are within three milliseconds from the angle from the camera frame, on average. Figure~\ref{fig:allgiment} (b) shows MARG and pressure values collected between two camera frames to correspond to a single frame. Finally, since the pressure is sampled at a lower frequency than the MARG sensor, Figure~\ref{fig:allgiment} (c) shows the MARG sensor reading closest to the corresponding pressure reading is kept, and the remaining samples in between the selected ones are discarded.

In this approach, small window sizes would only utilize signals corresponding to the selected camera frames. In contrast, overlapping with signals corresponding to previous frames is used to obtain more data for large window sizes.

After alignment, the final dataset contains five runs for each of the three object sizes. Each run consists of 900 camera frames, which had an average of 8 corresponding samples from the sensors per frame.
We used the data for all object sizes to ensure the dataset size was sufficient for model training.
Since all the sensors have different magnitudes and distributions, all the data apart from the object angle is scaled.
Finally, we implement standardization on the rest of the dataset. We use the following equation to normalize each sensor's data.

\begin{equation}\label{eq:stadardizing}
N^{(i)} = \frac{X^{(i)} - \mu^{(i)}}{\sigma^{(i)}}
\end{equation}

Where $N^{(i)}$ is the standardized signals of the $i^{th}$ sensor. $X^{(i)}$ are the raw signals of the $i^{th}$ sensor, $\mu^{(i)}$ is the mean signal value of the $i^{th}$ sensor. $\sigma^{(i)}$ is its signal's standard deviation.

\subsection{The angle estimation model}

Since tactile sensing measurements from objects under grasp manipulation are continuous and sequential, we used time series-based neural networks, specifically long- short-term memory (LSTM) based networks, to analyze windows sampling.

\subsubsection{Model Architecture}
Using a small baseline model initially, we arrived at the final model after adding layers that provided the best marginal improvement in performance for its size without overfitting, as increasing the model's size overfitted the training data.

\begin{figure}[thb!]
\centering
\includegraphics[width=\textwidth]{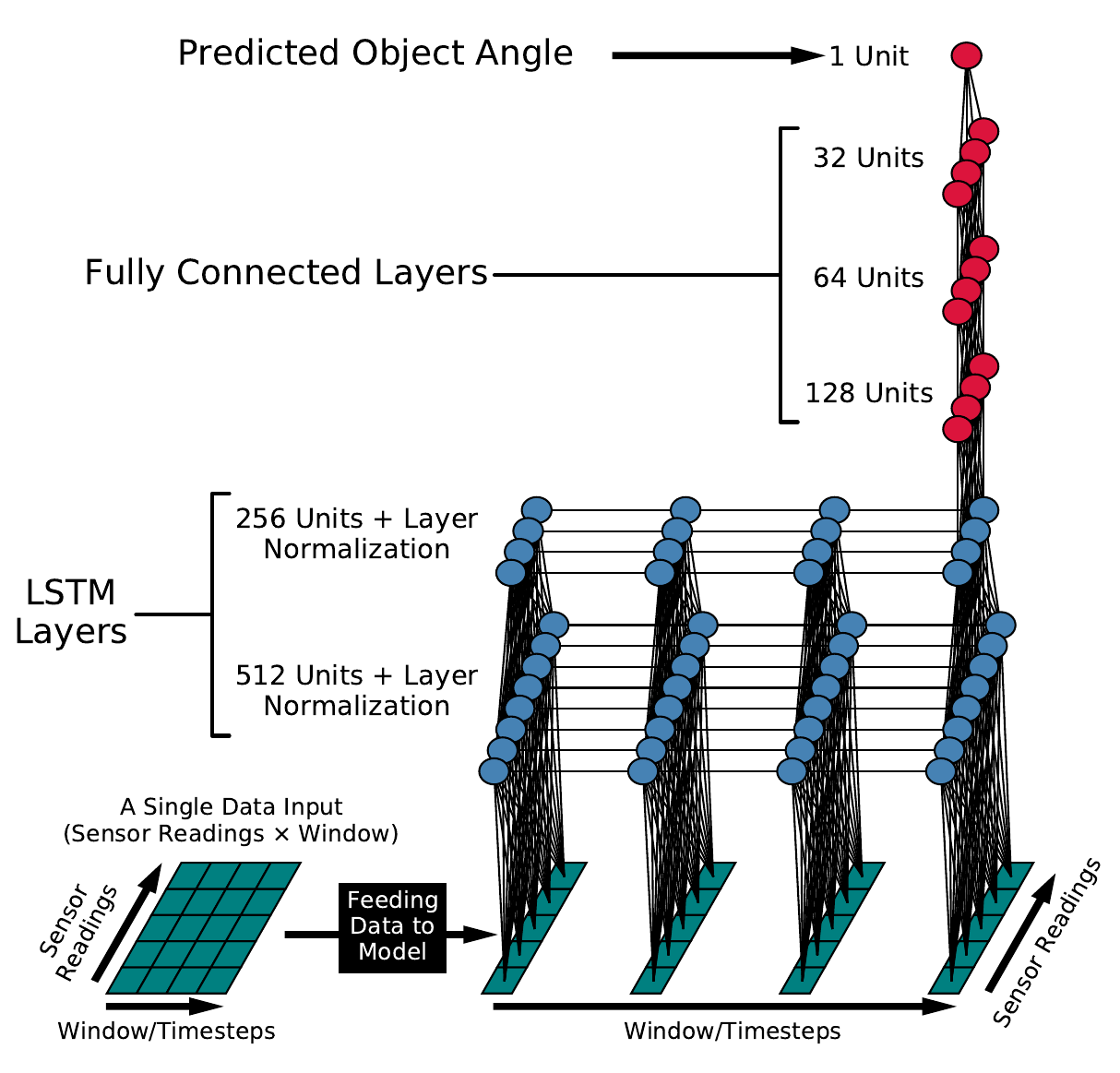}
\caption{The model architecture and the feed-forward of a single input sample through the model. The architecture consists of 2 LSTM layers with normalization layers with 512 Units and 256 units, respectively, and three fully-connected layers with 128, 64, and 32 neurons.}
\label{fig:arch}
\end{figure}

Figure \ref{fig:arch} shows the final model architecture we established consisting of 2 LSTM layers with normalization layers with 512 Units and 256 units, respectively, and three dense fully connected layers with 128, 64, and 32 neurons, respectively. 
All of the experiments are conducted on Compute Canada, an Advanced Research Computing platform, using the Python programming language and Tensorflow {\cite{tensorflow2015-whitepaper}} library to preprocess and model training.

We used the mean absolute error (MAE) between the angle's estimated and actual values as the training loss function. 
Moreover, we choose MAE as it diminishes in value much slower than a mean square error (MSE) as the model's estimation gets closer to the actual angle and has a value of less than one.

\subsubsection{Hyperparameters and Window Size Optimization}

Various experiments were performed to provide an understanding of the data and identify the effects of hyperparameters and performances corresponding to their variations. In particular, we manipulated batch sizes and windows and explored regularization methods. 
We explored the trade-off between window size and performance based on the best results to find the best gain in accuracy for a small model size.
This trade-off is fundamental in mobile robotics, with less memory and computational time leeway.
We performed a grid search to determine the hyperparameters over the number of epochs, learning rate, and batch size. We chose the best configuration of hyperparameters to conduct the study and investigate the window sampling technique. We used cross-validation with four folds, with six iterations for the model per fold on a rolling basis, to ensure the consistency of the model's performance, report any variance in the metrics scores and prevent data leakage.
Table \ref{tab:hyper} shows the neural network hyperparameters.

\begin{table}[thb!]
    \centering
    \caption{The hyperparameters' values of the neural network.}
    \label{tab:hyper}
    \begin{tabularx}{\textwidth}{cc}
    \toprule
    Hyperparameter  &       Value       \\
    \midrule
    Learning Rate   &       0.00025     \\
    Batch Size      &       128         \\
    Epochs          &       400         \\
    K-Folds         &       4           \\
    Iterations      &       6           \\
\bottomrule
\end{tabularx}
\end{table}

\section{Results}\label{sec:results}

Here we present the results of our experiment to estimate in-hand objects' orientation using a sliding window sampling strategy and evaluate with LSTM models.
The evaluation metrics used are mean squared error (MSE), mean absolute error (MAE), Coefficient of determination (R$^2$), and explained variance score (EXP).

\subsection{Model Training}

\begin{figure}[h]
    \centering
    \includegraphics[width=\textwidth]{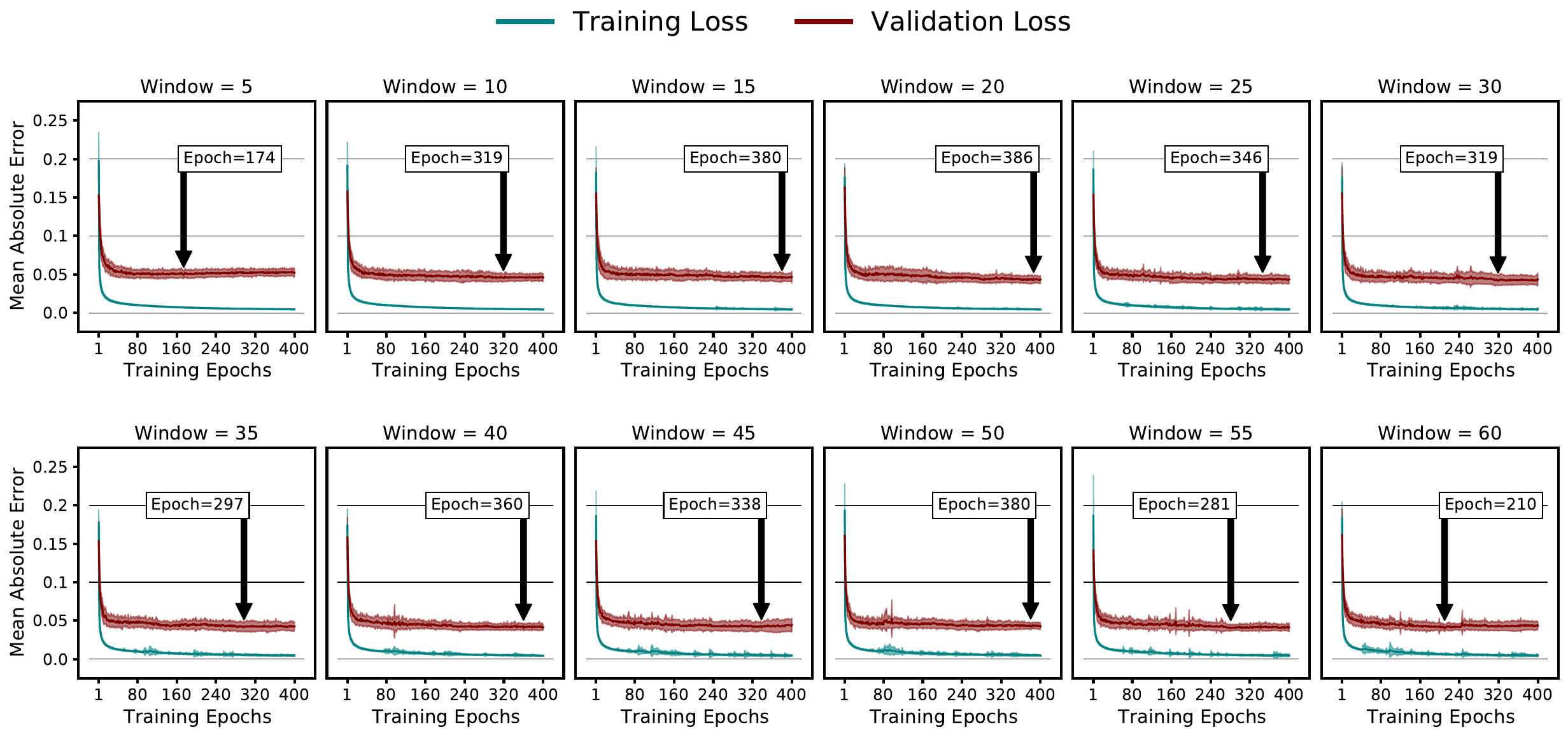}
    \caption{Average training and validation losses and their standard variation for the different window sizes, highlighting the average epoch of the lowest validation error.}
    \label{fig:training}
\end{figure}

Figure \ref{fig:training} depicts the training and validation losses during the training phase while highlighting the average epoch of the lowest validation error averaged over folds and model iterations.

We prevent overfitting by training the models for 400 epochs and selecting the model weights at the epoch of the lowest validation loss.

\subsection{Window Size}

\begin{figure}[thb!]
\centering
\includegraphics[width=\textwidth]{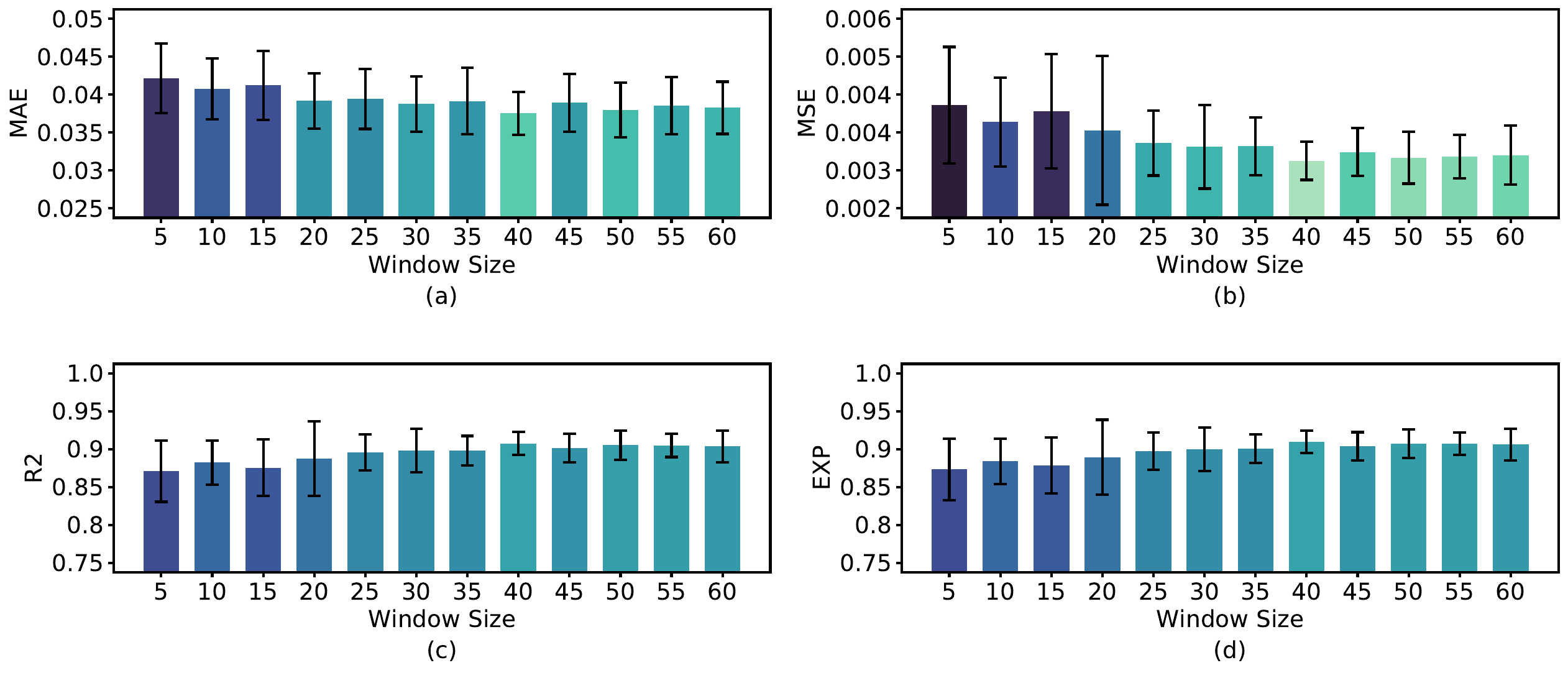}
\caption{The performance results from varying the window size. (a) Mean Absolute Error (MAE). (b) Mean Squared Error (MSE). (c) Coefficient of Determination (R$^2$ Score). (d) Explained Variance (EXP).}
\label{fig:test_metrics}
\end{figure}

The primary factor of temporal data explored was the window size. Figure \ref{fig:test_metrics} shows that a window size of 40 achieved the lowest error. It revealed a performance improvement as the window size expanded; however, the improvement magnitude decreased asymptotically.

The above result indicates that a window of 40 samples effectively captures the necessary amount of tactile information for estimating the object's orientation, regardless of the metric used.
Larger window sizes, beyond 40 samples, did not result in any further improvement in model performance. This finding is further supported by Table~\ref{tab:window_size}.


\begin{table}[thb!]
    \centering
    \caption{The detailed results of the inspected window size range using MAE and MSE errors in radian, R$^2$ score, and EXP.}
    \label{tab:window_size}
    \begin{tabularx}{\textwidth}{ccccc}
    \toprule
    Window &             MAE &             MSE &              R$^2$ &             EXP \\
    \midrule
     5 & 0.0422 ± 0.0046 & 0.0042 ± 0.0012 & 0.8710 ± 0.0404 & 0.8732 ± 0.0408 \\
    10 & 0.0408 ± 0.0040 & 0.0038 ± 0.0009 & 0.8823 ± 0.0292 & 0.8840 ± 0.0297 \\
    15 & 0.0412 ± 0.0046 & 0.0041 ± 0.0012 & 0.8754 ± 0.0374 & 0.8785 ± 0.0370 \\
    20 & 0.0392 ± 0.0036 & 0.0036 ± 0.0016 & 0.8873 ± 0.0492 & 0.8894 ± 0.0493 \\
    25 & 0.0394 ± 0.0040 & 0.0034 ± 0.0007 & 0.8956 ± 0.0237 & 0.8975 ± 0.0246 \\
    30 & 0.0388 ± 0.0037 & 0.0033 ± 0.0009 & 0.8981 ± 0.0287 & 0.8997 ± 0.0289 \\
    35 & 0.0392 ± 0.0044 & 0.0033 ± 0.0006 & 0.8981 ± 0.0193 & 0.9005 ± 0.0188 \\
    40 & \color{blue} 0.0375 ± 0.0028 & \color{blue} 0.0030 ± 0.0004 & \color{blue} 0.9074 ± 0.0153 & \color{blue} 0.9094 ± 0.0148 \\
    45 & 0.0389 ± 0.0038 & 0.0032 ± 0.0005 & 0.9013 ± 0.0190 & 0.9038 ± 0.0185 \\
    50 & 0.0380 ± 0.0036 & 0.0031 ± 0.0005 & 0.9053 ± 0.0192 & 0.9069 ± 0.0188 \\
    55 & 0.0385 ± 0.0037 & 0.0031 ± 0.0005 & 0.9048 ± 0.0153 & 0.9073 ± 0.0149 \\
    60 & 0.0383 ± 0.0034 & 0.0031 ± 0.0006 & 0.9037 ± 0.0208 & 0.9060 ± 0.0207 \\
\bottomrule
\end{tabularx}
\end{table}

The model achieved the best MAE of 0.0375 radians with a window of 40 and an average error of 0.0408 with a window size as small as 10 samples. The model also obtained high R$^2$ and EXP scores of 0.9074 and 0.9094, respectively for the best window size.

\begin{figure}[ht!]
\centering
\includegraphics[width=\textwidth]{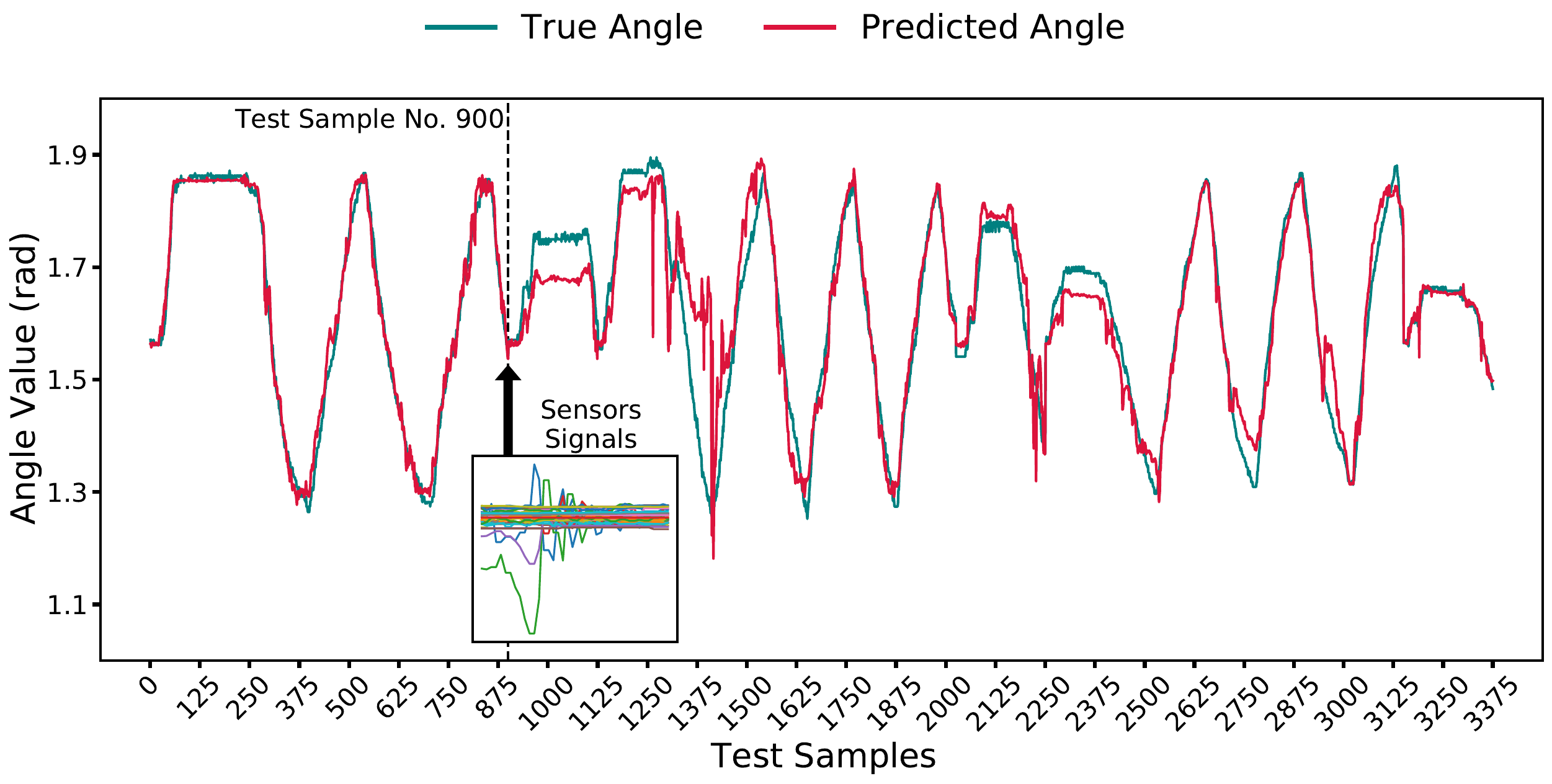}
\caption{A comparison of the predicted and ground truth angle for one of the iterations of the model with the best window size of 40 samples. A 40 samples window corresponding to the prediction of angle sample 900 is highlighted. There are persistent deviations on some peaks and troughs such as around sample 1000, 1250 and 2375 which are more likely to occur when the object is static. This may occur because no new information is obtained by the sensor when the object is static and as a result the error in prediction can't be corrected.}
\label{fig:prediction}
\end{figure}

We use one of the iterations of the best model to illustrate its angle prediction compared to ground truth in Figure \ref{fig:prediction}.
The figure also shows a window of 40 samples of sensors' readings that correspond to a single angle prediction, test point no. 900.

\subsection{Comparing this Temporal Deep Learning Method to Ridge Regression}

 Nevertheless, we trained linear and ridge regression models with the same data protocol we applied for the neural networks for comparison.
Notably, these two classifiers present the best results in an approach that does not use the time-series relation in the data \cite{PradodaFonseca2019}, in which the models were trained per object size and not using all the sizes at once. 
Although we can not conclude the advantage of temporal data from the MAE and MSE values due to different normalization and scaling procedures ranges, the R$^2$ and EXP scores highlight that point in the previous study \cite{PradodaFonseca2019}.
The results of these two models are reported in Table~\ref{tab:regression} using our preprocessing procedure for comparison.

\begin{table}[ht!]
    \centering
    \caption{The results of standard regression models using MAE and MSE errors in radian, R$^2$ score, and EXP.}
    \label{tab:regression}
    \begin{tabularx}{\textwidth}{ccccc}
    \toprule
    Model &             MAE &             MSE &              R$^2$ &             EXP \\
    \midrule
    Ridge Regressor &  0.0677 &  0.0088 &  0.6875 &  0.7033\\
    Linear Regressor &  0.0678 &  0.0089 &  0.6862 &  0.7021\\
\bottomrule
\end{tabularx}
\end{table}

\section{Discussion}

This study aimed to determine if pose estimations relate to the time series tactile data captured by the sliding window sampling strategy adopted. We analyzed the performance of using a neural network with LSTM layers in estimating the angle of the handled object by tactile sensing robotic hand, considering different sliding window sizes of input samples.
The deep learning model is compared to standard regression models to showcase the improvement due to their temporal tactile data incorporation.

We presented a data processing procedure to align collected data from multiple asynchronous sensors and approximate their reading timestamps to yield multi-sensor temporal data in Figure {\ref{fig:allgiment}}. The data was then used to train and evaluate a deep learning model that we optimized its architecture, as shown in Figure {\ref{fig:arch}}, and training hyperparameters using grid search.

By testing a range of window sizes between 5 and 60 to investigate the degree of the impact of the temporal relation between tactile data, we demonstrated the importance of such relations between sensor readings in estimating the angle of an object under grasp.
We found that incorporating a small window size of 5 inputs gives an acceptable performance of 0.0422 radians, equivalent to 2.417 degrees, and scores above 0.87 for both R$^2$ and EXP metrics.
Compared to the standard classifiers tested in this study, we found that the smallest window can improve about 26\% and 24\% for the R$^2$ and EXP scores, respectively, and a reduction of 0.0256 and 0.0047 for MAE and MSE, respectively. Thus, it shows that the temporal relationships of the sensor readings can improve estimating the objects' angle as evident in Tables \ref{tab:window_size} and \ref{tab:regression}.

Furthermore, these results gradually improved by integrating more sensor information from larger window sizes of up to 40 samples per window, after which the performance saturates.
Including more past readings beyond 40 samples did not add valuable information to the instantaneous angle value prediction as seen in Figure \ref{fig:test_metrics}.
This result shows that despite the importance of temporal relationships in tactile data for estimating the object's angle during manipulation, these relationships diminish asymptotically after a threshold.

For the best window size of 40 samples, we found that it achieves an acceptable error for many applications with an average of 0.0375 MAE in radians and can explain most of the variance in the distribution, shown in 0.9074 R$^2$ score and 0.9094 EXP score.
This performance is sufficient in multiple applications without a camera reference during the grasp phase, thus supporting the use of temporal tactile data for orientation estimation of in-hand objects in unstructured environments.

Notably, the model can achieve such results after training on data from objects with differing sizes, thus incorporating more variation in the data, making the temporal relation harder to capture. Therefore, improving on the previous results \cite{PradodaFonseca2019} where only a per object angle estimation was performed. Additionally, using different object sizes also generalizes the model performance. This generalization also extends to being applied in an under-actuated system which experiences larger effects of parasitic motions (compared to fully actuated systems). However, this is a limited application that does not account for the other dimensions, and, as a result, future work can include all other axis and provide a complete object pose description and improve the robustness.
In addition, we can not directly compare the metrics because of different normalization methods, as they use a normalized degree unit, whereas we use radians.

Future research can use our results as a reference and investigate a tactile dataset with objects of different shapes as well as remaining degrees of freedom to determine the complete change in the object's pose, not only its yaw orientation.
Moreover, feature engineering can be an additional step alongside the temporal tactile data to enhance the model further. Future studies can benefit from the proposed alignment of asynchronous sensors that we illustrated in Figure {\ref{fig:allgiment}}.

Finally, collecting a dataset of both arm-mounted and gripper-mounted tactile data for object orientation estimation can further illustrate the benefits of temporal tactile sensing compared to other techniques.

\section{Conclusions}

This paper illustrates the importance of temporal tactile data in estimating the orientation of in-hand objects by proposing a model architecture with LSTM layers that uses signals from tactile sensors on the fingers.
We evaluated these experiments' performance using MAE, MSE, R$^2$, and the EXP metrics. The results show that including temporal data benefits the orientation estimation of the objects up to an asymptotic threshold, as investigating a range of window sizes concluded that the smallest window studied boosts the performers compared to standard regression models, such as linear and ridge regression.
The best window size in the investigated range is 40 input samples, which could predict the object angle with an average MAE of 0.0375 radians.
Our model also has an R$^2$ value of 0.9074 and an EXP value of 0.9074, respectively. By comparison, the ridge regressor yields an average MAE of 0.0677 radians, 0.6875 R$^2$ score, and 0.7033 EXP value.
Therefore, the relationship between the tactile signals object's angle is better explained with time-series models that utilize the temporal relationships of the sensors' readings. These results highlight the benefits of using previous state information, particularly because manipulation tends to be sequential. At the same time, it presents a simple architecture that uses less processing and computational power compared to setups with high-density tactile sensors. Moreover, our tactile data model can work with objects such as symmetric cylinders that may look fixed from the visual sensors' perspective. Finally, it also presents the viability of pose estimation without needing 3d models.

Our proposed model can be included in future research investigating the pose estimation problem using tactile data and the importance of their temporal relations with different modes of pose change. Future studies can also benefit from our proposed preprocessing procedure to match the timestamps of readings obtained from asynchronous sensors.
\chapter{Grasp Approach Under Positional Uncertainty Using Compliant Tactile Sensing Modules and Reinforcement Learning}
\label{chap:ch4_pick_objects}
\section{Introduction}

Grasping objects is a fundamental yet intricate task in robotics applications ranging from industrial automation to assistive technologies.
Traditional approaches to robotic grasping have predominantly relied on visual sensory information to discern the environment and guide the manipulation strategies \cite{Mathew2018}.
However, visual perception is susceptible to challenges such as variable lighting conditions, object transparency, occlusion, camera calibration errors, and clutter, despite its rich detail. These factors can significantly impede the robot's ability to accurately estimate an object's pose, especially within unstructured environments \cite{Hao2023, Hoda2016, jiang2022shall}.

\begin{figure}[!t]
    \centering
    \includegraphics[width=\columnwidth]{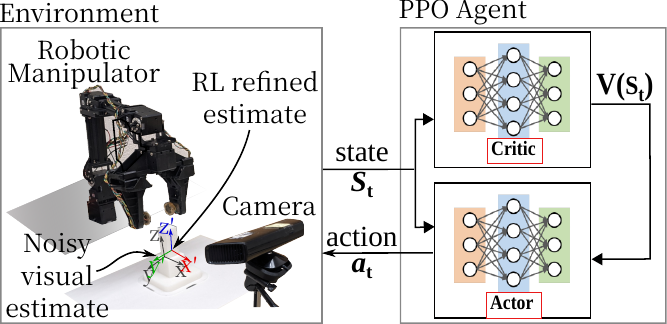}
    \caption{The camera provides an estimate of the object's location. The manipulator attempts to grasp it and return the system's state, i.e., the success/failure, the force experienced by each sensor and the orientation of the sensor to the reinforcement learning agent. The agent uses that information to update the position to attempt a grasp again if unsuccessful prior.}
    \label{fig:system}
\end{figure}

Tactile sensing may offer a way to circumvent some of the limitations associated with visual perception. Tactile sensors in robotic tasks have been used in various capacities, such as for pose estimation \cite{Galaiya2023,prado2019estimating}, texture classification \cite{Lima2021,lima2020dynamic}, and object recognition \cite{Pohtongkam2023,da2022tactile}. Additionally, a wide breadth of features can be collected from them, including contact force, contact location, and contact deformation \cite{Mandil2023}. Compliant tactile sensors can deform upon contact and provide direct feedback on the robot's interaction with its environment. This form of feedback is invaluable, especially when visual information is unreliable or insufficient.

Moreover, the potential of compliant tactile sensors providing nuanced feedback can be amplified when integrated with perception capabilities empowered by techniques like reinforcement learning (RL). RL can enable robots to learn from interactions with their environment, refining their performance through trial and error. This learning paradigm promises to significantly enhance the efficiency and reliability of robotic grasping in the face of positional uncertainties.

\begin{figure*}[!ht]
    \centering
        \includegraphics[width=\columnwidth]{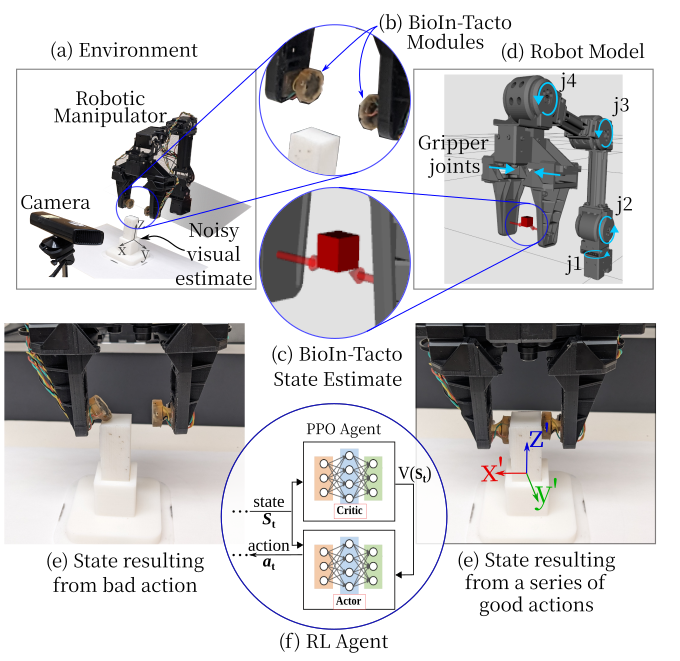}
    \caption{The blue arrows are the direction of the joint movement of the manipulator. The Environment provides the orientation of the sensors attached to the end effector and the pressure experienced by the sensors. This information is sent to the agent every time a grasp is attempted. The agent sends a new position for trial if unsuccessful. }
    \label{fig:environment}
    \vspace{-0.5cm}
\end{figure*}

Our preliminary findings demonstrate the potential of our method to significantly reduce the number of attempts required to achieve a successful grasp, highlighting the efficacy of integrating tactile feedback with reinforcement learning for adaptive grasping under uncertainty. We do this by using an estimate of the object position from sources such as cameras and use reinforcement learning to explore the vicinity of the estimate to identify and grasp the object based on the reward determined by the tactile information as shown in Fig.~\ref{fig:system}. 

\section{Related Work}

Recent research has focused on addressing this challenge by investigating feature sensing and robotic grasping of objects with uncertain information \cite{wang2020feature}.
Significant work has been done to improve tactile robotic grasping using reinforcement learning. 
Most methods consist only of vision-based sensing methodologies that use supervised learning, reinforcement learning, or a combination of both \cite{Kleeberger2020}.
Also, methods like domain randomization \cite{James2017} have reduced the gap from simulation to reality, although learning in simulation did not generally transfer better to reality.

Matak \textit{et al.} \cite{matak2022planning} used a vision and tactile system for grasp planning and execution for a multi-fingered robotic hand. 
The tactile sensors are used to determine if the predicted contact matches and, if not, to modify the grasp. 
Dong \textit{et al}. \cite{dong2021tactile, dong2019tactile} use tactile sensing and reinforcement learning to insert objects, the second phase of the peg-in-the-hole task in which they explored only one dimension.
They use the Gelsim \cite{gelslim} tactile sensor to obtain surface contact information and a force/torque sensor on a parallel jaw gripper as a baseline.
Jiang \textit{et al}. \cite{jiang2022shall} devised a framework for grasping transparent objects using visual and tactile sensing. 
They trained a neural network to segment the image and determine the region to touch using the GelSight tactile sensor.

Due to their deformable nature, compliant tactile sensors can dynamically probe objects' surfaces, providing rich tactile feedback that can reduce uncertainty. 
Alves de Oliveira \textit{et al.} \cite{Oliveira2017, de2023bioin} proposed the BioIn-Tactom, a compliant bio-inspired tactile sensing module that integrates various sensing modalities to provide rich tactile feedback for robotic systems.
Da Fonseca \textit{et al.} \cite{da2022tactile} addressed the challenge of tactile object recognition during the early phases of grasping using underactuated robotic hands, later proposing an approach that enables robots to perform manipulation tasks more effectively by inferring object orientation from tactile feedback \cite{prado2019estimating}.

Welyhorsky \textit{et al.} \cite{welyhorsky2022neuro} introduced a neuro-fuzzy grasp control system for a teleoperated robotic hand, enhancing dexterity and precision during grasping tasks. Zhu \textit{et al.} \cite{zhu2020teleoperated} investigated teleoperated grasping using a robotic hand and a haptic-feedback data glove, improving the operator's sense of touch and control over the robotic hand during manipulation tasks.


This paper aims to develop a pre-grasp approach that mitigates positional uncertainty using compliant tactile sensing modules and reinforcement learning techniques.
Our approach is motivated by the need to reduce uncertainty in object perception and execute successful grasps even when the object's properties are ambiguous or unknown. 
\section{Methodology}

To evaluate the use of RL for contact exploration, the present work used an environment setup that consists of one 4-DoF OpenMANIPULATOR-X (model RM-X52-TNM) equipped with two BioIn-Tacto sensing modules mounted on the gripper fingers (see Fig.\ref{fig:manipulator}).

\subsection{Compliant Tactile Sensing}

\begin{figure}[!t]
\centering
    \begin{minipage}[t]{0.55
    \columnwidth}
        \centering
    \includegraphics[width=1\columnwidth]{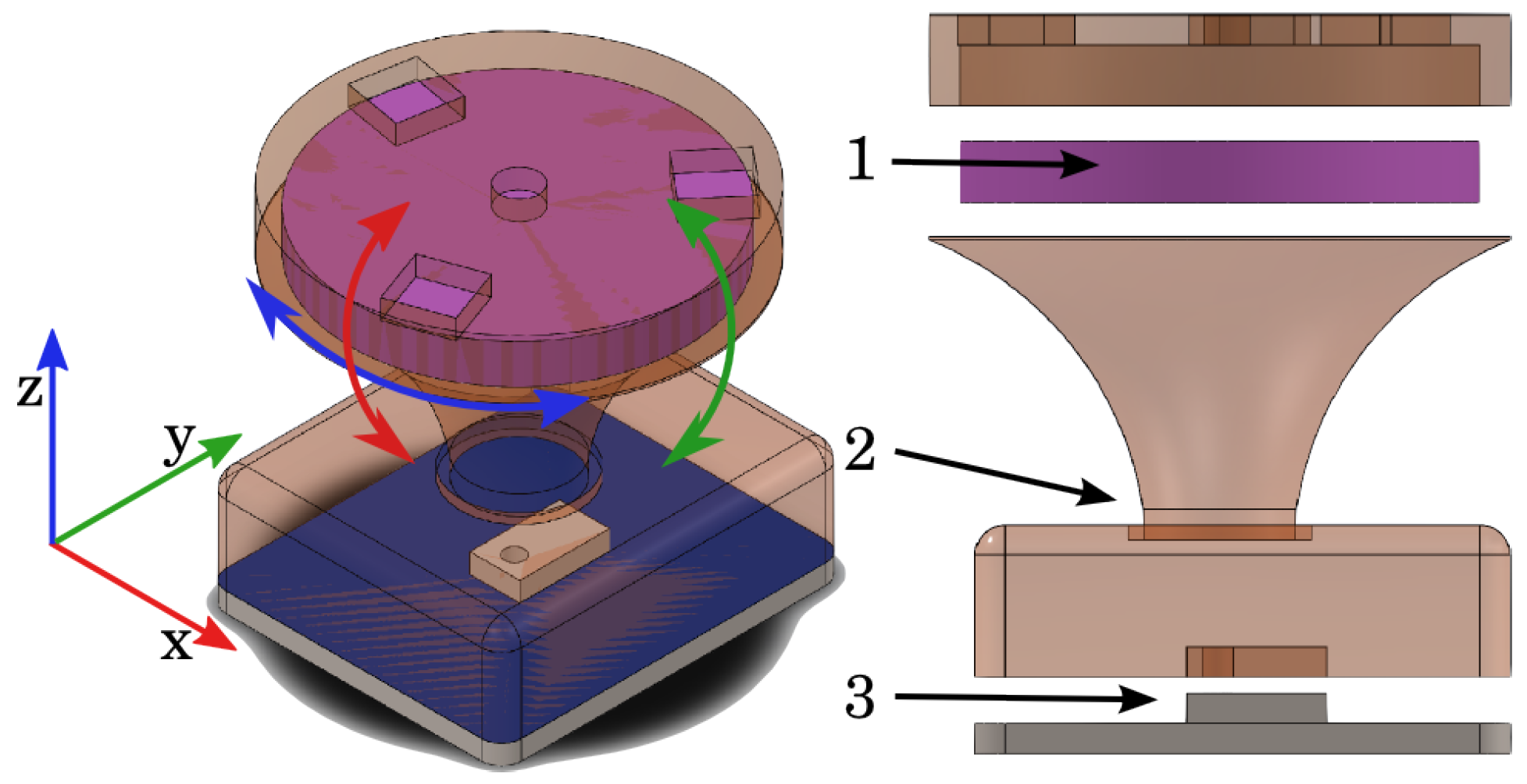}
    \end{minipage}
    \hspace{2mm}
    \begin{minipage}[t]{0.35\columnwidth}
        \centering
    \includegraphics[width=0.8\columnwidth]{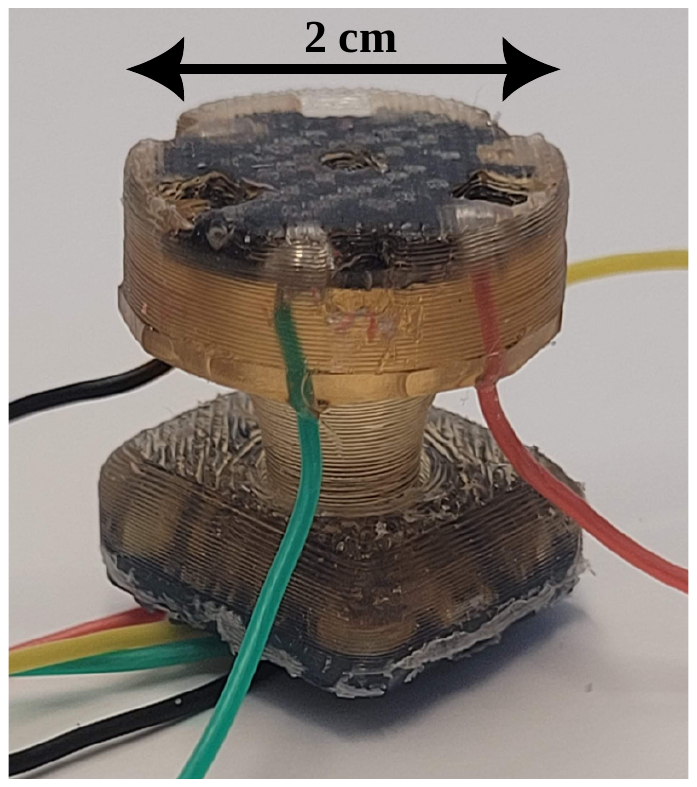}
    \end{minipage}
    \caption{Tactile sensing module. Components: 1—MARG (magnetic, angular rate, and gravity) system; 2—compliant structure; 3—barometer \cite{Oliveira2017,
prado2019estimating, de2023bioin}.}
    \label{fig:sensing_mod}
\end{figure}

The pose uncertainty underscores the necessity for employing a compliant tactile sensor.
To perform the experiments required, we mounted two BioIn-Tacto modules on each gripper's jaws. Fig.~\ref{fig:sensing_mod} shows the BioIn-Tacto sensor used in this setup.

The tactile sensors comprise an inertial measurement unit (IMU) and a barometer. 
We obtained the orientation of the sensor with respect to the globe using the accelerometer and the gyroscope, which is passed through the Madgwick filter. 

Let: $\mathbf{a}$: acceleration vector, $\mathbf{g}$: gyroscope vector, $\mathbf{q}$: quaternion representing orientation, $\beta$: filter gain, $\Delta t$: time interval between sensor updates.
Then, the Madgwick filter update equations are:

\begin{equation}
\begin{aligned}
\dot{\mathbf{q}} & = \frac{1}{2}\mathbf{q} \otimes \begin{bmatrix} 0 \\ \mathbf{g} \end{bmatrix} - \frac{\beta}{2}\mathbf{P}\mathbf{S}(\mathbf{q}^*)\mathbf{J}(\mathbf{q})^T\mathbf{a}, \\
\mathbf{q} & = \mathbf{q} + \dot{\mathbf{q}} \Delta t.
\end{aligned}
\label{eq:medwick}
\end{equation}

where $\otimes$ denotes quaternion multiplication, $\mathbf{P}$ is a $4 \times 4$ identity matrix, and $\mathbf{S}$ and $\mathbf{J}$ are functions to compute the skew-symmetric matrix and Jacobian matrix, respectively. By integrating these equations, the Madgwick filter provides accurate and stable orientation estimates, enabling precise pose estimation even in the presence of noise and uncertainty.
This is then transformed to obtain the sensor's orientation relative to the manipulator's base frame.

The pressure obtained is normal to the surface of the sensor. 
Since the sensor's surface either conforms to the object's surface, it grasps, or is distorted by a collision, the pressure determined by the sensor is proportional to the distortion it experiences in the normal direction to the surface. 
This pressure is normalized to the initial reading obtained during the home position so that the observation space is of the same magnitude.  

The MARG sensor data and manipulator inverse kinematics determine the real-time contact normal orientation and position.
The sensing module's flexible structure provides a frame for the estimation of contact orientation and position at each contact tentative without requiring a planning approach.

\subsection{Reinforcement Learning}

The present paper uses an exploratory approach, leveraging the compliant sensor structure to evaluate approach attempts and contact.
The robot assumes an initial grasp position that has a level of uncertainty (see Section \ref{sec:uncertainty}).
Every each approach tentative in case of contact a RL strategy is updated. 

The robotic agent controls this process and chooses actions implementing a Proximal Policy Optimization (PPO) as presented in Algorithm \ref{alg:iteration}. The network consists of 2 fully connected layers with 64 units each.

In the algorithm, each approach tentative starts with the $reset()$ function, taking the agent back to an initial state, which is set to be above the target object at a position $p_{i}$ given by the uncertainty model $\mathbb{U}$.

The initial observations are passed to the $step()$ function, which starts looping while the $done$ signal is set to False.
Once the done signal changes to True, the episode ends, sequences of states and cumulative rewards are stored, and a new episode can begin. 
Once $n_{steps}$ reaches $total\ timesteps$, the iteration is completed, and the policy is updated using PPO.

Each training iteration consists of several episodes, which vary depending on how soon the agent reaches a terminal state.
We choose to work with 2048 steps per training iteration, meaning that the policy will be updated after 2048 steps have been executed over the number of episodes necessary.

\begin{algorithm}
  \caption{Aproach tentative iteration}
  \label{alg:iteration}
  \begin{algorithmic}[1]
    \Require $obs; p_{i} \in \mathbb{U}; total\_timesteps$.
    \State $n \gets 0$
    \State $done = False$
    \While{$n_{steps} \leq total\_timesteps$}\Comment{Iterate episodes.}
    \State $obs_{ini} \gets Reset(\mathbb{U})$
    \State $obs_{t} \gets obs_{ini}$
    \While{done = False} \Comment{Iterate steps.}
    \State $obs_{t+1}, r_{t+1}, done \gets Step(obs_{t})$
    \State $ n_{steps} \gets n_{steps}+1$
    \EndWhile
    \EndWhile
    \State Update policy with PPO.
  \end{algorithmic}
\end{algorithm}

Algorithm \ref{alg:iteration} is defined around five main elements: the action and observation spaces, the $reset()$ and $step()$ methods, and the reward function. 
The \textit{action} space contains all actions that the agent can execute.
In this case, our action is a single value $a \in [-1;1]$ for the base actuator z-axis.

The observation space $obs$ is composed of joint efforts $effs \in \mathbb{R}^4$ in $N.m$, barometer level $baro \in \mathbb{N}$, and orientation $orient \in \mathbb{R}^3$.
It describes the environment to the agent and is used to compute actions.

The $reset()$ method is responsible for taking the agent back to an initial valid state where the agent receives the initial observations $obs_{ini}$ from the sensing module.
In the present work, the robot is performing a fine adjustment of the approach position, which implies that the manipulator is above the target object.
This means that, during this phase, it first retreats to an initial pose called $home$, making no contact with the object.
A point $p_{i}$ is sampled from the uncertainty model $\mathbb{U}$, and the manipulator moves to the given location, hovering above the object.
Then, the initial observations $obs_{ini}$ are sent to the $step()$ function.

The $step()$ method, as shown in Algorithm~\ref{alg:step}, is responsible for sending the state $S_{t}$ to the policy through the observations $obs_{t}$ and getting and action $a_{t}$ back, executing it and calculating the reward $r_{t}$ based on the resulting observations. 
It is also responsible for verifying that the agent has reached its goal and ending the training episode.
We condition the end of the episode to happen under two scenarios: 1) the number of steps for an episode reaches $n_{steps}=50$ or 2) the barometer level and the MARG pose variations indicate an excess of contact pressure or wrong contact (see Eq.~\ref{eq:reward}).
This allows the manipulator to keep changing the initial position $p_{i}$ at the end of every episode for generalization purposes and also avoid catastrophic contact with the target object.

\begin{algorithm}
  \caption{Step}
  \label{alg:step}
  \begin{algorithmic}[1]
    \Require $obs$; $p_{i} \in \mathbb{U}$, where $ p_{i} = \theta_z$
    \State  $orient, effs, baro \gets obs$\Comment{Real-time input.}
    \State $a \gets Policy(obs)$\Comment{$a \in [-1;1]$}
    \State $pos_{new,i} \gets p_{i} + orient\times step \times a$ \Comment{$step=0.005m$}
    \State Move manipulator to $pos_{new}$\Comment{Eq.~\ref{eq:pos_new}.}
    \State $reward \gets f(sensors)$. \Comment{Eq.~\ref{eq:reward}.}
    \State Checks if the terminal state has been reached.
  \end{algorithmic}
\end{algorithm}

Equation~\ref{eq:reward} shows the reward function $f(sensors)$ is a function of the MARG and barometer levels where (a) if there was a significant change in the orientation angle of the sensors (indicating a collision); (b) negative if there was no grasp; (c) a smaller magnitude of negative reward if there was too much pressure from one sensor but not from the other since it was close to grasp but offset; (d) finally, positive on a successful grasp where the object can be lifted and placed without any slippage.

\begin{equation}
        f(sensors) =
    \begin{cases}
    -0.1 & \text{if } orien > threshold \text{ (a)}\\
    -0.1 & \text{if } baro < threshold \text{ (b)}\\
    -0.1 & baro1 > thres \oplus baro2 > thres \text{ (c)}\\
    +0.5 & baro1 \oplus baro2 > threshold \text{ (d)}
    \end{cases}
    \label{eq:reward}
\end{equation}

\subsection{Object Pose Uncertainty\label{sec:uncertainty}}

We introduced an error term to the measurement to simulate the noise inherent in the pose estimation of an object, specifically along the x-axis.
This error term follows a normal distribution characterized by its mean ($\mu$) and standard deviation ($\sigma$). The normal distribution equation for the x-axis error is given by:

\begin{equation}
f(x|\mu, \sigma) = \frac{1}{\sigma \sqrt{2\pi}} e^{-\frac{(x-\mu)^2}{2\sigma^2}}
    \label{eq:pos_new}
\end{equation}
where:
$f(x|\mu, \sigma)$ is the probability density function,
$\mu$ is the mean of the distribution along the x-axis,
$\sigma$ is the standard deviation along the x-axis, and
$x$ is the variable of interest along the x-axis.

By incorporating this error term into our measurements along the x-axis, we are able to mimic a $0.02m$ variability and uncertainty present in real-world pose estimation scenarios.

\begin{figure}[!htbp]
    \centering
    \includegraphics[width=\columnwidth]
    {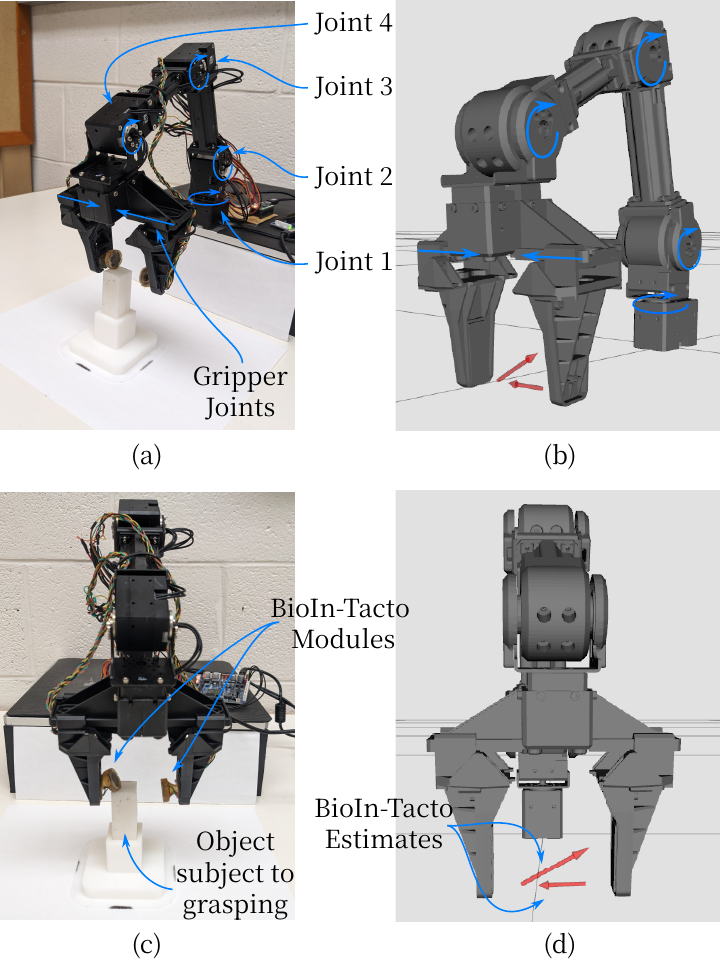}

    \caption{The manipulator attempts to grasp the white cuboid by exploring the area within the vicinity of the estimated location provided. }
    \label{fig:manipulator}
\end{figure}

We used a four-degree-of-freedom robotic manipulator with two sets of sensors attached on each side of the end effector and a 3D printed cuboid for the object to be grasped, as shown in Fig~\ref{fig:manipulator}.

\section{Experimental Setup}

\begin{figure*}[t!]
    \centering
   \includegraphics[width=\columnwidth]{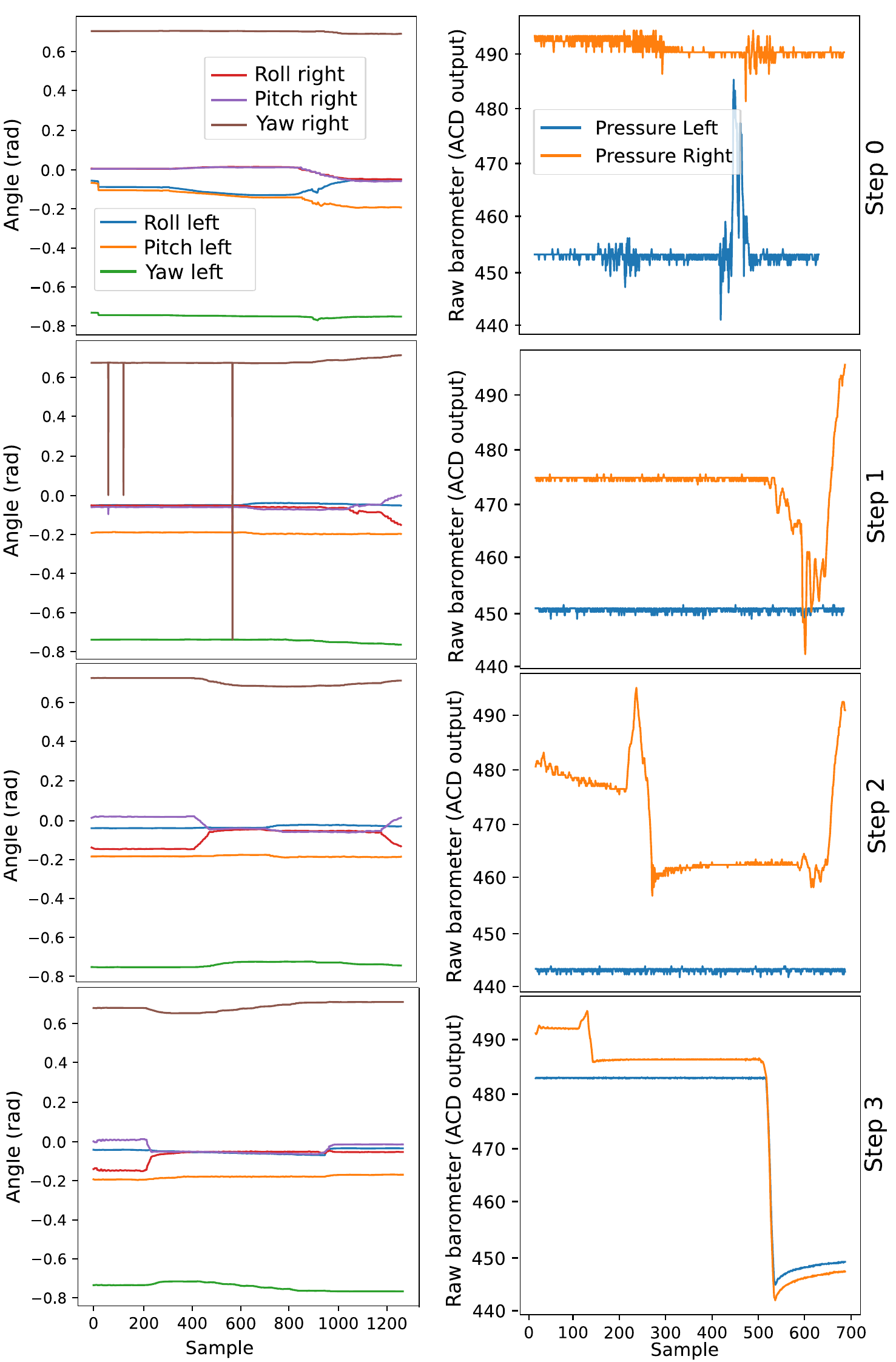}
    \caption{The pressure and the orientation of each sensor on the end effector during each step in the episode.}
    \label{fig:sensoroutput}
\end{figure*}

The environment, that is, the pressure experienced by the and orientation of the sensors as well as the pseudo-position of the object to be grasped are provided to the reinforcement learning agent, the agent then provides the manipulator the estimated position of the object. Subsequently the results of the attempted grasp based on the estimate are provided to the agent. If successful, the environment is reset. Conversely, if unsuccessful the agent tries to estimate again and the cycle repeats as shown in Algorithm \ref{algo:graspexploration}.

Fig. \ref{fig:sensoroutput} shows a graph of each sensor output during each step on an episode. The first row is the attempt at trying to grasp the object but doesn't interact with it at all. For steps 1 and 2, the right sensor experienced a collision with the object and that provided feedback that could be used to correct the orientation. Finally on step 3, the object is grasped successfully. 

\begin{algorithm}

\caption{Grasp exploration}\label{euclid}
\label{algo:graspexploration}

\begin{algorithmic}[1]
\For{each episode
}:
    \State Move manipulator to the home position
    \For{each step
    }:
        \State attempt grasp using position from agent
        \State calculate reward 
        \State update policy
    \EndFor
    \If{successful grasp
    }:
    \State end episode
    \EndIf
\EndFor
\end{algorithmic}
\end{algorithm}

\begin{figure*}[!t]
    \centering
    
    \includegraphics[width=\columnwidth]{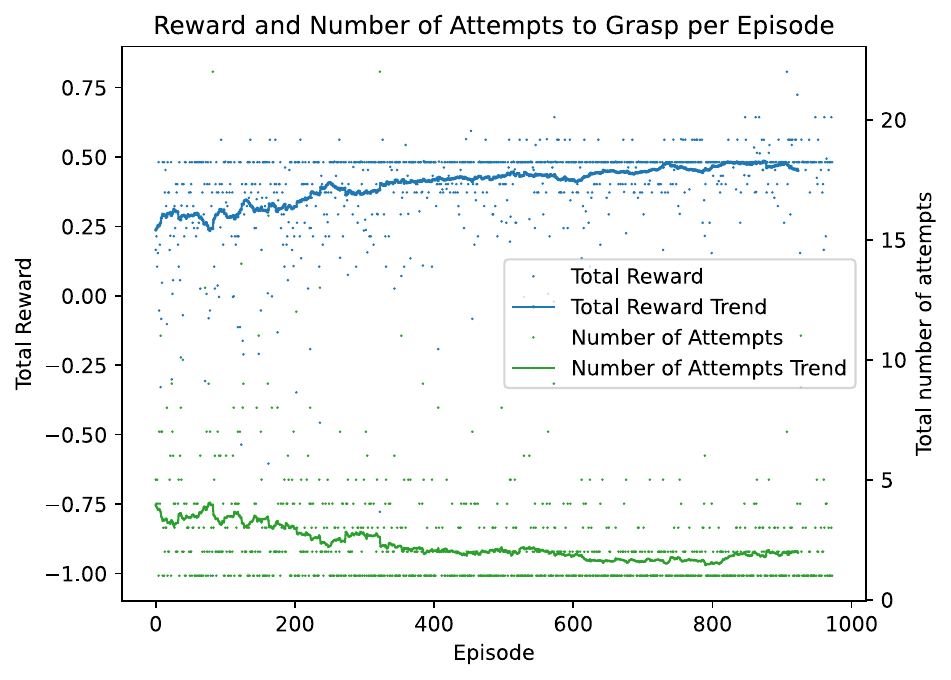}
    \caption{The trend line is calculated with a 50 moving average. The trend line of the reward increasing with the episodes and the number of steps per episode reducing. As the number of episodes increase the rewards and number of steps reach an asymptote. }
    \label{fig:Reward&steps}
\end{figure*}
\newpage

\section{Results}

The results presented here show our evaluation of the learning rate when the grasp approach is affected by uncertainty.
Fig.~\ref{fig:sensoroutput} shows an example of the agent learning behavior of our system, which corresponds to a positive reward in the last step in which there was a successful grasp and negative rewards for the rest.

The agent learning effect can also be observed in Fig.~\ref{fig:Reward&steps}, which shows the total reward increasing while proportioning the number of attempts to grasp reducing successfully. 
The 50-episode moving average trend shows that as the model iterates, it could grasp the object with fewer attempts and, consequently, in a shorter time span on average despite the number of attempts varying from one attempt to the next. 
Although more than 20 steps were taken at some points, most of the episodes of high density occurred at the start, when the model had the fewest iterations. 

However, we still start at an average of 5 steps, and it drops to an average of close to 2 steps. 
As the number of episodes increases, the density of the number of steps is mainly below the 5-step line, showing that the system improves the search pattern, minimizing the number of attempts to grasp.
\section{Conclusion and Future Work}

Our investigation introduces a method that leverages compliant tactile sensors and reinforcement learning to enhance the robotic grasping approach under object pose uncertainty. 
The results underscore the substantial potential of this approach, demonstrating a remarkable improvement in grasp accuracy and efficiency and a notable reduction in the number of attempts needed for a successful grasp.
By fusing tactile feedback with adaptive learning, we significantly advance robotic capabilities, paving the way for systems that can more effectively comprehend and interact with their environment.
Future work will involve applying the method to various objects with different materials, shapes, and scenarios to assess its versatility and effectiveness.

\chapter{Human Example Pretraining Influences on Reinforcement Learning for Obscured Object Trajectory Guidance }
\label{chap:ch5_pull_objects}
\section{Introduction}


With the widespread use of robotics in assembly lines, the push for more general-purpose robots has become reinvigorated over the last decade. However, there are open questions that need to be addressed to substitute human efficacy in tasks like unscrewing, plugging and unplugging cables, and machine disassembly for repair in unstructured environments. In particular, understanding the environment and performing as well as humans do. The human ability to use the multitude of senses together or as needed provides much-needed flexibility in trajectory identification of visually obscured objects. Aspects of this challenge could be reduced to the peg-in-hole problem, which has been a widely researched assembly task \cite{beltran2020variable, unten2022peg, nigro2020peg}.  


Insertion and extraction of peg-in-hole is a stepping stone to a variety of manipulation tasks that require local exploration of the environment. Especially for fixed cameras where the targets are occluded by other objects. This exploration provides information on object interaction boundaries between the target and objects in the environment. Tactile information can take the reigns in such instances. Many works exploring the peg-in-hole problem tend to use some form of contact sensing capability ranging from force/torque \cite{Jin2021, SONG2021101996} to vision \cite{kim2022}. Apart from giving information that the camera cannot obtain from occlusion, it can also provide information that is not available using a camera, and this information can be leveraged to make localized corrections. This is especially true for compliant grippers for which the dynamics of the hand can not be predicted based on the known geometry and kinematics distorted by unknown object properties. 

Sensors such as e-skin \cite{Xiong2022}, visiflex \cite{Fernandez2021}, GelSight Fin Ray \cite{Liu2022}, and BioIn-Tacto \cite{de2023bioin} are examples of the foray into compliment tactile sensors that are being developed for the characteristics of the human fingertips that can conform to the shape of the object being grasped and also give local information not distorted by the conformation. These developments benefit manipulation tasks that rely on fine motor movements with environmental uncertainty to determine contact states. These contact states can guide reinforcement learning algorithms \cite{wang2021alignment}.

Lu \textit{et al.} \cite{LU2022} presents a human-centric model of manufacturing in which humans work with emphatic machines in a symbiotic relationship, with one of its fundamental components being human-machine understanding and, in particular, understanding the instruction, action and goal. One of the challenges in explaining the process of achieving the goal is its impreciseness with the variability of the environment and the task. One method of reducing impreciseness while reducing the training space is by human examples. In this section, we explore the effects of pretraining the machine learning model using human-based imperfect examples in speeding up the ability of the agent to infer tactile signals and consequently speed up the success in object extraction in peg-in-hole tasks using reinforcement learning. Viability in this domain can give reason to explore similar techniques in applications like prosthesis control, where pretraining is commonly done for everyday prosthesis use.  
\section{Literature Review}
As robots take over more human tasks, the need for effective human-robot knowledge transfer increases. Lee \textit{et al.} \cite{LEE2024} developed a human-robot knowledge transfer framework. Humans and robots have different modes of perception, communication and reasoning. Therefore, knowledge has to be transferred using learning, comprehension and translation. For dis/assembly tasks, they categorize assembly knowledge into assembly design, plan, and action. They suggested that an agent can consider a demonstration as a knowledge transfer using the assembly plan. 

Using interactions of the manipulator with the object and the object with the environment is one strategy used to extrapolate local information and consequently increase the likelihood of a successful peg-in-hole insertion as done by Kim et al. \cite{kim2022}. They used a vision-based compliment tactile sensor attached to the phalanges to develop a reinforcement policy based on the estimate of the misalignment of the first attempt. The tactile sensor provides a deformation image that is used to determine the deformation angle of the sensor based on a trained convolution neural network. They use the tactile inferred angle and proprioception as factors that they explore to determine the contact line of the object and the hole boundary. This is used to determine the following action for insertion. They use a tactile feedback controller to move the robot down and then rock or pivot the object during the exploration phase to determine the contact line. More than 95\% of their attempts are successful in all six object and hole shape combinations in which the clearance of the object and hole was an average of 2.25mm. 

Dong \textit{et al}. \cite{dong2021tactile} explore methods to align the object and the hole in peg-in-holes tasks without using prior knowledge of the object or the shape or contact locations. They compare supervised and reinforcement learning, curriculum training, and the tactile representation of a light-based tactile sensor, comparing tactile flow to RGB representation. They determined the RGB sensors with tactile flow representation and curriculum training reinforcement learning policy performed the best compared to either when everything else was constant and 
\begin{itemize}
    \item force/torque is used instead of a vision-based tactile sensor,
    \item curriculum training is not used,
    \item RGB representation is used instead of tactile flow.
\end{itemize}

Simulation to real-life transfer is another commonly used practice used to leverage the generally needed high amount of episodes required for reinforcement learning \cite{lammle2022, Kozlovsky2022, Beltran2020, de2019end, lima2020dynamic}. Xie \textit{et al}. \cite{xie2023learning} use simulation to real-life transfer learning with human demonstration for new objects to reduce the transfer cost. They use an RGB camera to segment the peg and the hole using human guidance to create masks to train a segmentation network. Subsequently, they use a deep neural network to determine the position and rotation features of the segmentation masks. Finally, they use an Advantage Actor-Critic network to create an insertion policy based on the rotation and position features. They take into account the features of occlusion using the Long Short-Term Memory network to account for the history that is otherwise limited by a single frame during occlusion. They obtained an accuracy of greater than 90\% in both, simulation and real-world applications for hole tolerances less than 1mm. They addressed the occlusion problem by considering the time series of the frames. The best result for unseen objects was 65.6\%, which they suggest remains a big challenge, especially in dynamic environments.  

This challenge of occlusion is significantly amplified in object extraction since there is no initial bearing to extrapolate from either. Therefore, we explore using tactile sensing to determine the trajectory the manipulator needs to follow to remove the peg entirely from a hole. 
\section{Materials and Methods}

\begin{figure}[!ht]
    \centering
    \includegraphics[width=0.9\columnwidth]{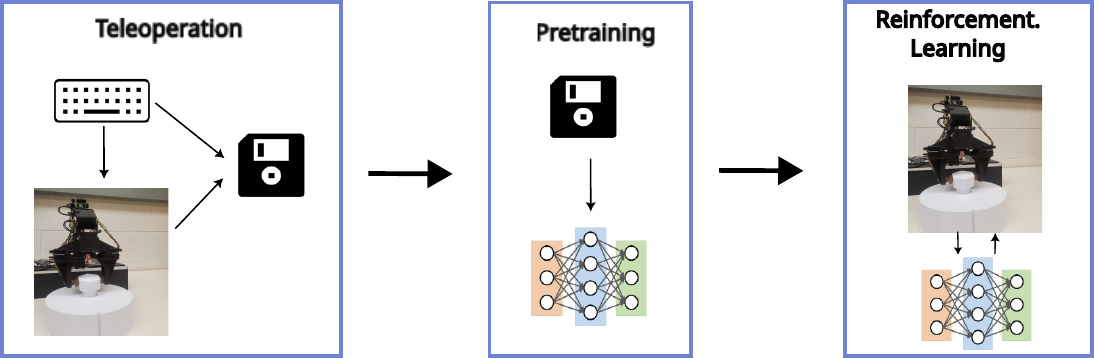}
    \caption{The process is broken down into three steps. We first record the actions and the environment using teleoperation for object extraction in the peg-in-hole task. We use the recorded data to train a neural network. We use the trained neural network as a foundation for the Q-learning reinforcement learning setup for the object extraction. }
    \label{fig:pull_overview}
\end{figure}

We use a four-degree-of-freedom manipulator with two tactile sensors attached to each phalanx to perform object extraction from a peg-in-hole setting where most of the peg is occluded. The method involves three stages, as shown in figure \ref{fig:pull_overview}. First, we perform teleoperation to perform the peg-in-hole object extraction task, recording the inputs and behavior. Next, we train a neural network based on the data collected during teleoperation. Finally, we use a reinforcement learning agent with the trained neural network to perform object extraction.

\subsection{Sensor and Manipulator Setup}

\begin{figure}
    \centering
    \hspace*{0cm}\includegraphics[width=0.6\columnwidth]
    {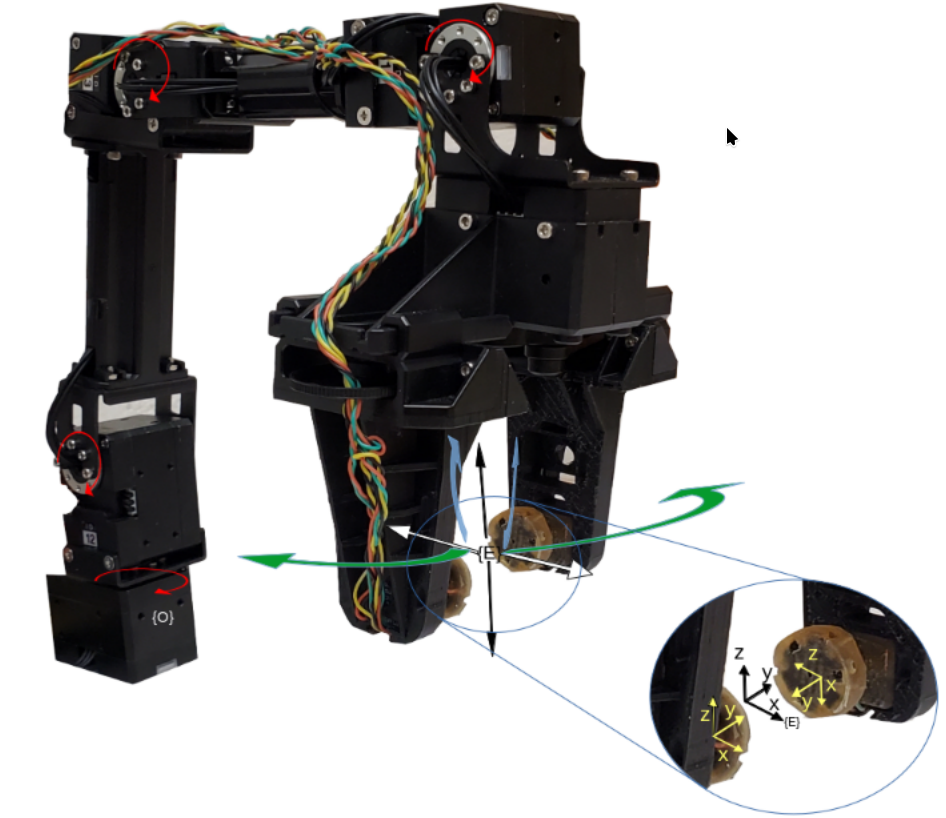}
    \caption{This is the manipulator and sensor setup we are using for this experiment. The four degrees of freedom manipulator with two tactile sensors, one attached at the end of each phalanx. The red arrows represent the joint motions for the four degrees of freedom manipulator. The manipulator is controlled with respect to the end effector frame $\mathbb{E}$ in the directions shown with the black, green, blue and white arrows. The positions with respect to the base frames $\mathbb{O}$ are converted to the positions with respect to the end-effector frame $\mathbb{E}$. Resulting in motions on the x, y and z axis as well as rotation about the y and z axis with respect to $\mathbb{E}$. We also obtained the position of the tactile with respect to $\mathbb{O}$ sensor based on the data it provided.}
    \label{fig:manipulator_teleop_arrows}
\end{figure}

We use a four-degree-of-freedom robotic manipulator with the BioIn-Tacto compliant tactile sensor attached to each phalanx of the manipulator as shown in \ref{fig:manipulator_teleop_arrows}. Due to the four-degree freedom limitation, the manipulator could only move at any point in the XZ and YZ plane but only on a curved trajectory in the XY plane with respect to the end-effector frame while maintaining the orientation. The manipulator can be controlled linearly in the X and Z directions and can rotate in the Y and Z axis of the end effector frame $\mathbb{E}$. If the new end effector pose $\mathbb{F}$, is the current end-effector pose that shifted in only in either the individual X, Y, or Z directions or rotated only in the Y or Z direction.  

The tactile sensor provides magnetic field, angular velocity, linear acceleration in the three dimensions, and pressure perpendicular to the surface of the sensor. 
We obtain the orientation of the sensor using the angular velocity and linear acceleration.

We obtain the orientation using the Madgewick filter using  equation \ref{eq:medwick}.
We use the scaled pressure, the end effector pose and the change in orientation in quaternion for each tactile sensor. The pressure reading is scaled in a range of $0 - 1$ from $0-400$ by dividing by $400$, and the change in orientation is  
 obtained using     
 \begin{equation}
     \mathbf{Q_{\Delta}} = \mathbf{Q_{t}} \otimes  \mathbf{Q_{t-1}^{-1}} 
 \end{equation}

  Where $\otimes$ denotes quaternion multiplication, $\mathbf{Q_{t-1}} $ is the orientation of the sensor at $t-1$ in quaternion, $\mathbf{Q_{t}}$ is the orientation of the sensor at $t-1$ in quaternion and $\mathbf{Q_{\Delta}}$ represents the change in orientation between time $t$ and $t-1$ to prevent the distortion of error accumulation from the integrator since the most relevant information is based only on the recent information. 

\newpage
\subsection{3D printed Object Set used for Extraction}

\begin{figure}[!h]
    \centering
    \includegraphics[width=0.6\columnwidth]{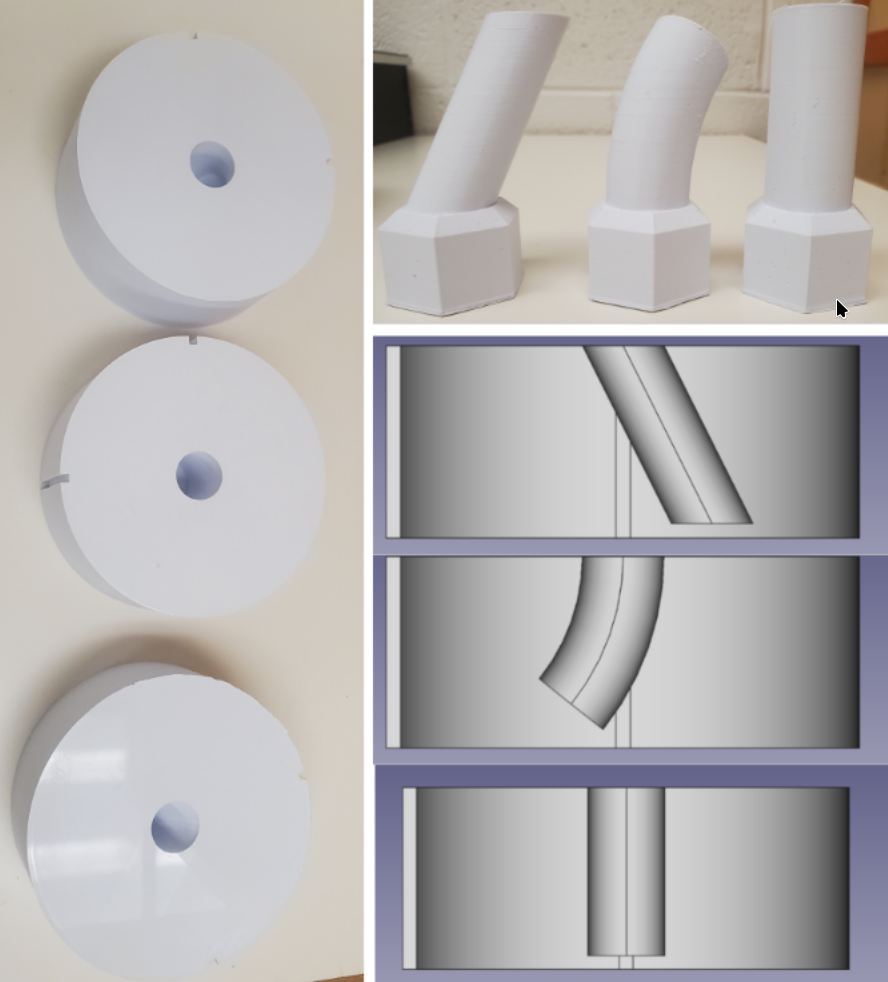}
    \caption{These are pegs and the respective holes used in this peg-in-hole setup. \textbf{Top right:} a slanted, curved and vertical peg. \textbf{Left:} the respective holes for each of the pegs. \textbf{Bottom right}: the cross-section of respective holes. }
    \label{fig:pull_objects}
\end{figure}

We 3D printed the three objects for the pulling task with varying levels of complexity to compare the cross-compatibility of training. A vertical peg, a slanted peg and a curved peg with its respective holes as shown in figure \ref{fig:pull_objects}.

\subsection{Teleoperation and Data collection for Pretraining}
The first step is collecting data using teleoperation. Here, we configure the manipulator to be controlled by a user using the keyboard. Peg one was attempted to be picked three times, since it is symmetric. Whereas pegs two and three are picked three times in each orientation of $0^\circ$, $45^\circ$, $90^\circ$ and $135^\circ$. The user pressed the keys on the keyboard corresponding to the direction in which they wanted to move the end-effector to guide the peg out of the hole. The corresponding input, the end-effector pose, the pressure on each tactile sensor, and the orientation of each tactile sensor are recorded. 

\subsection{Pretraining the Neural Network}

The user keyboard inputs are encoded using a one-hot encoder. We use a three dense layer deep neural network and a final softmax layer. Each layer contained 64, 32, and 16 units, respectively. It was trained based on the data collected using teleoperation. We explore performance on no pretraining, pretraining with the same object, training with a different object, training with the other two objects and pretraining with all three objects.  

The pertaining is used to reduce the amount of time it takes for the early sessions of reinforcement learning sessions. Since it is governed by the randomness of the initial Q-network and the learning rate $\alpha$, the discount factor $\gamma$ and the window of exploration required based on the action space in the bellman equation:
\begin{equation}
Q(s, a) = (1 - \epsilon)Q(s, a) + \alpha \left[ r + \gamma \max_{a'} Q(s', a') - Q(s, a) \right]
\end{equation}
Where $s$ is the state, $a$ is the action, $\epsilon$ is the exploration parameter, $r$ is the immediate reward, $s'$ is the next state, and $a'$ is the next action. 

 \subsection{Object extraction using Reinforcement Learning}
 
 We use the deep Q-learning reinforcement learning algorithm \cite{mnih2013playing} due to the large and continuous state space. All teleoperation actions took less than thirty-two steps to complete, so a batch size of thirty-two steps was chosen. The network is updated every four steps since the task can be completed in less than thirty-two steps based on the teleportation, and similarly updating the target network is every eight steps. The discount factor was 0.75 since the previous information was not very significant. We increased the exploration capability by setting the epsilon greedy parameter to 200 for the session in which we did not use any pretrained network so that the agent could explore more, whereas, for the pre-trained network, we used the greedy parameter of 50. 

Equation~\ref{eq:pull_reward} shows that the reward $reward$ is a function of the barometer levels and height of the end-effector where (a) if there was a large change in pressure indicating there was a lot of friction experienced from the walls of the base if there was a large change in the pressure of the sensors (indicating a collision); (b) a function of the change in height of the end effector if the friction was not large; (c) finally, positive reaching the designated height.

\begin{equation}
        reward =
    \begin{cases}
    -0.5 & \text{if } threshold\text{ }pressure < baros  \text{ (a)}\\
    f(\Delta end-effector\text{ }height) & \text{if } baros < threshold\text{ }pressure \text{ (b)}\\
    +1 & end-effector\text{ if } height >  threshold \text{ }height\text{ (c)}
    \end{cases}
    \label{eq:pull_reward}
\end{equation}

\section{Results}
The results shown here explore the effects of tactile information used for object extraction in the peg-in-hole task for which the training phase can be sped up with imperfect human examples. In particular, whether pretraining reduces the training episodes needed, the effects of training time on pretraining using other objects, and the effects of pretraining on the complexity of the trajectory required. 

\begin{figure}[!h]
    \centering
    \includegraphics[width=0.9\columnwidth]{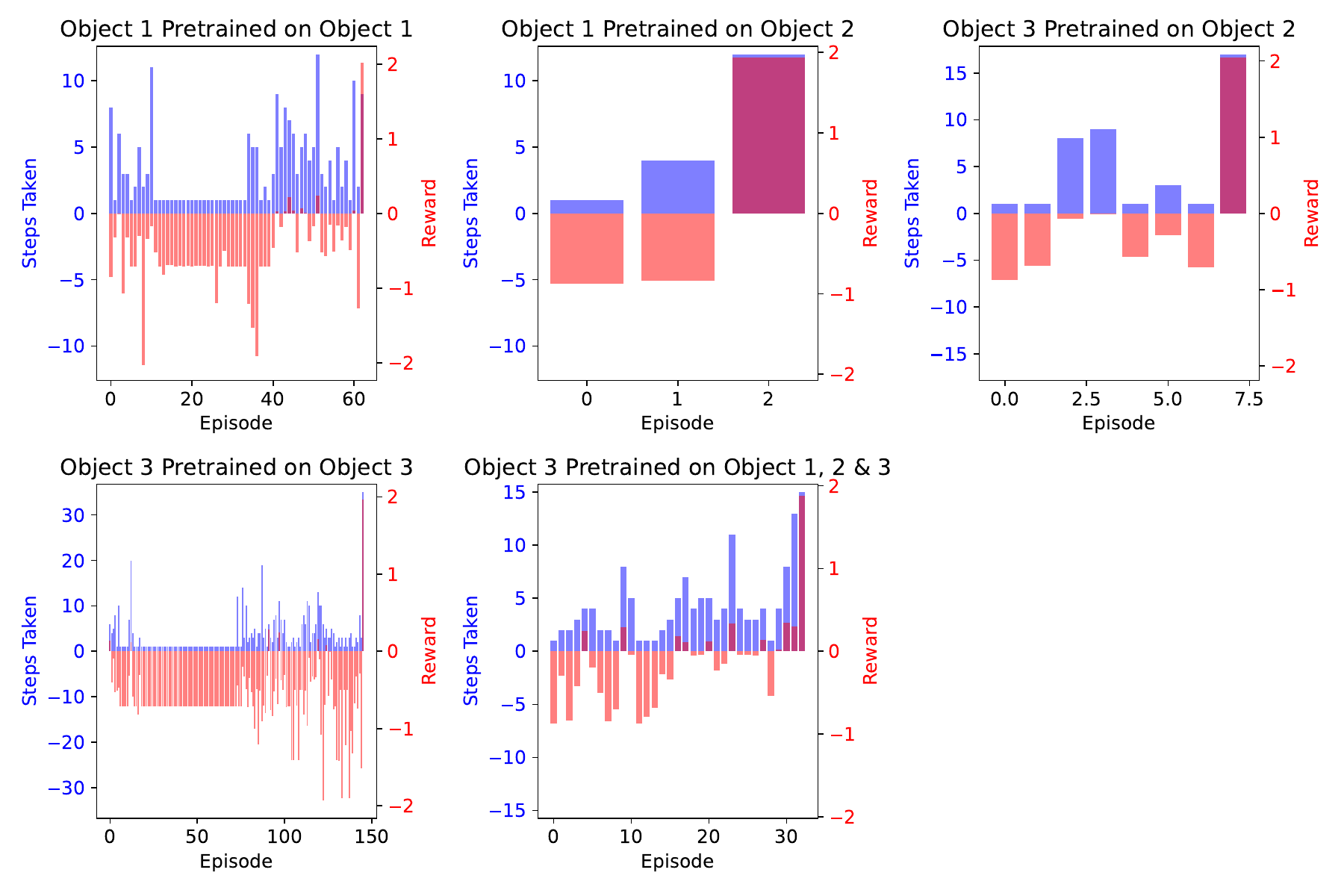}
    \caption{This figure shows the reward that corresponds to the steps for each episode for some trials until the last episode, where the attempt was successful. Object one pretrained on object one and took more than 60 episodes.}
    \label{fig:steps_and_rewards}
\end{figure}

Figure \ref{fig:steps_and_rewards} shows a variety of behaviour of the manipulator based on the environment, object, and the pretraining methodology. Object one, pre-trained on object two, took only three episodes, whereas object three, pretraining on object three, took more than 150 episodes, with a series of episodes having the same number of steps and rewards when the updated network has an equally bad next step. 

\begin{figure}
    \centering
    \includegraphics[width=0.7\columnwidth]{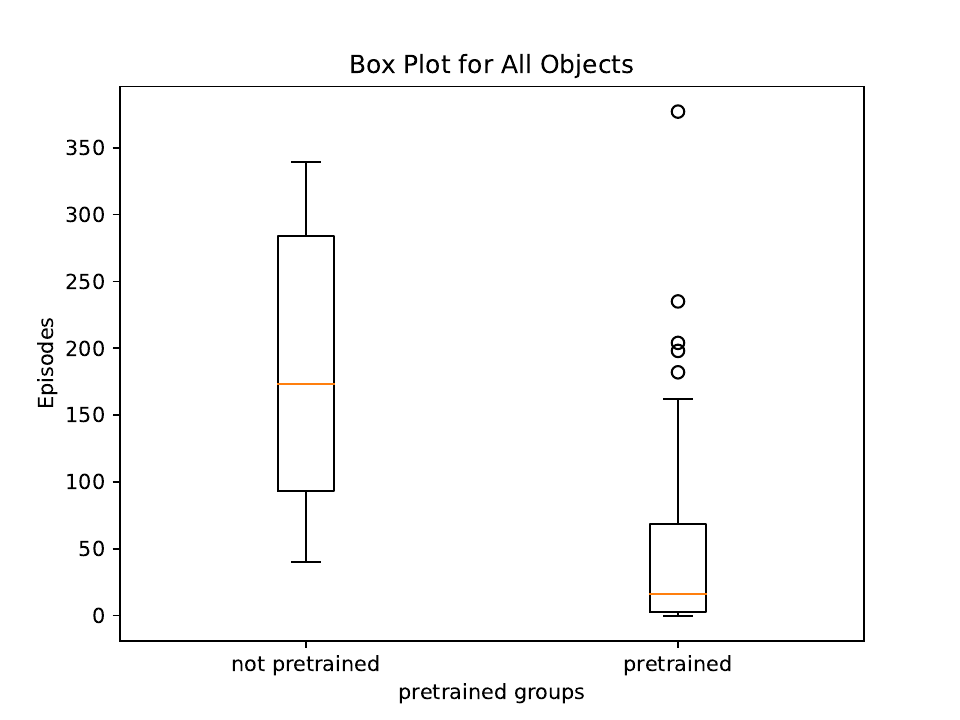}
    \caption{This figure shows a comparison between the number of episodes required to complete an extraction on all objects with pretraining or no pretraining.  }
    \label{fig:all_pretrained_vs_not}
\end{figure}

The agent successfully learnt to extract all three pegs using just tactile information. Without pretraining, it took an average of 178 episodes to complete the extraction of the object. This is substantially improved upon as shown in Figure \ref{fig:all_pretrained_vs_not} with pertaining. It took an average of 48 episodes with pretraining on all objects. Significantly fewer episodes were required to obtain a successful extraction after pretraining. Some outliers can be attributed to the oscillation from fewer random motions which may result from a local trough that is cyclical.

\begin{figure}[!h]
    \centering
    \includegraphics[width=0.9\columnwidth]{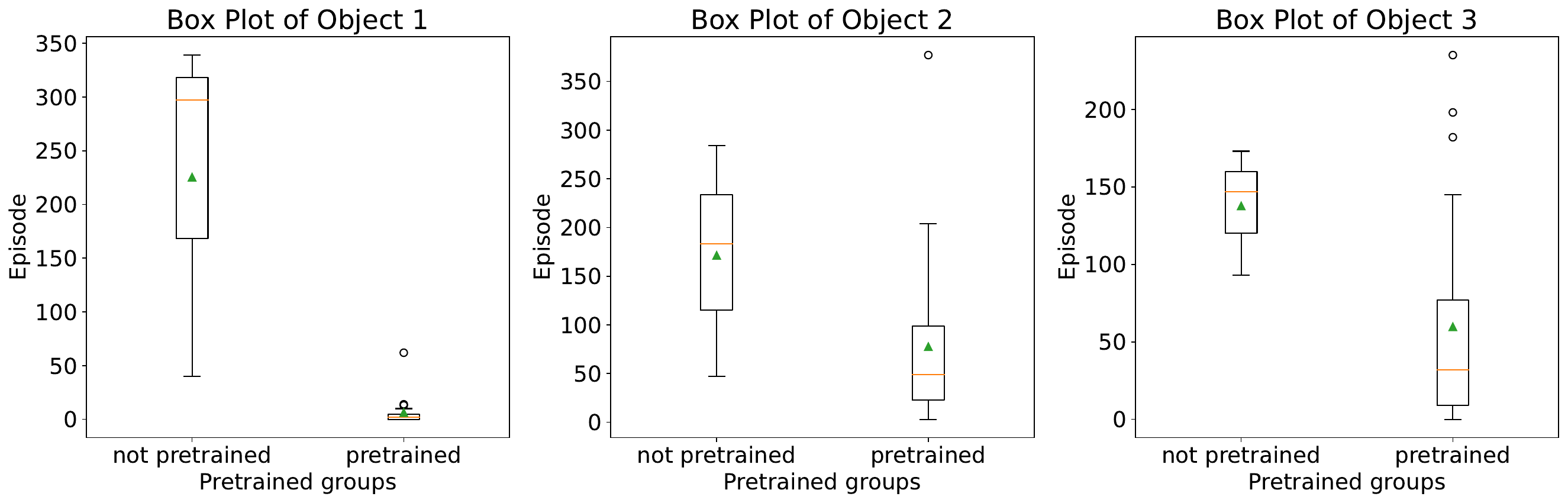}
    \caption{This figure shows a comparison between box plots of the number of episodes taken for a successful extraction with any form of pretraining and no pretraining for each object.}
    \label{fig:per_object_pretrained_vs_not}
\end{figure}

Figure \ref{fig:per_object_pretrained_vs_not} shows that the effects of pretraining on all three objects were substantial. However, for the complicated objects, there were some attempts that required more episodes after pretraining compared to no pretraining. This could be explained by the fact that pretraining on another object and reducing the exploratory phase results in a longer recovery period from learning the wrong actions. 

\begin{figure}[!h]
    \centering
    \includegraphics[width=0.7\columnwidth]{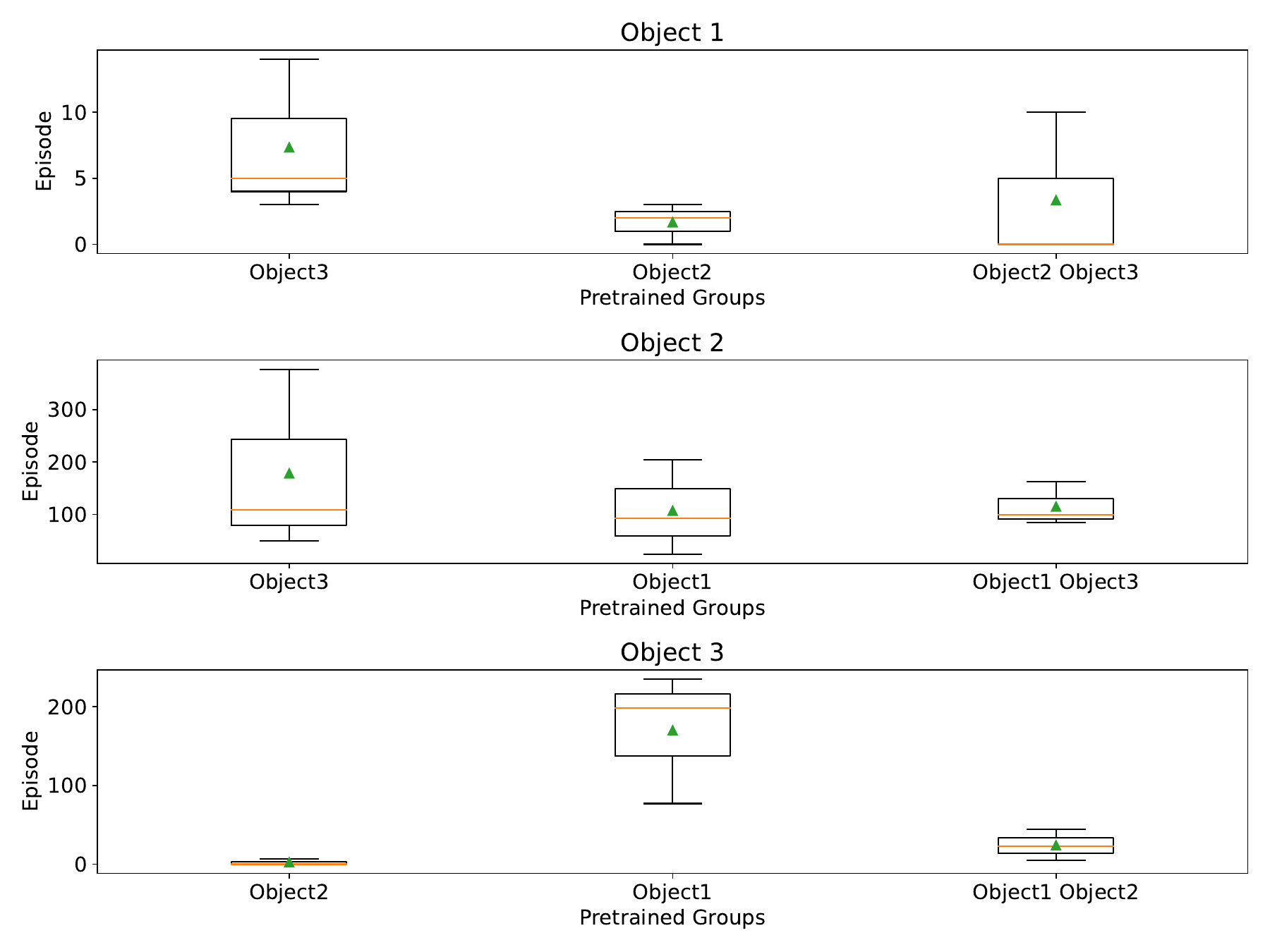}
    \caption{This figure shows box plots for the number of episodes required to successfully extract the object based on the pretraining on other objects. This explores the generalizability of the pretraining. }
    \label{fig:pretrained_on_other_objects}
\end{figure}

Figure \ref{fig:pretrained_on_other_objects} shows the effects of pretraining on different objects. For object one, the type of pretraining did not affect the number of episodes as much as it did for objects two and three. Since object one was simpler than objects two and three, the training in any of the objects may have captured the behaviour required for a successful episode more easily. For object two, pretraining on objects one and three together had a smaller range of the total number of episodes required to obtain a successful extraction. However, pretraining on object 3 had a worse average and upper bound, this could be because there was less transferability from object two knowledge compared to object one and may have distorted the behaviour corresponding to the training. On the other hand, pretraining on object two performed substantially better than pretraining on object one to extract object three.

\section{Conclusion and Future works}

In general, pretraining resulted in fewer episodes being required to successfully extract an object. However, the type of pretraining also influenced the number of episodes. In particular, more complicated trajectories were sensitive to the limits of pretraining on a simple trajectory. Moreover, the pretraining on more complicated objects assisted the simpler objects to be extracted faster. 

Future works should explore more complicated extraction objects to determine the extent of pretraining and to identify the dimension of pretraining required to cover a large subset of instances that can be experienced by the manipulator. 

A holistic system should also be explored for insertion and extraction tasks combined to determine the crossover between the two tasks and if prior knowledge in insertion will assist in extraction. 










\chapter{Conclusion and Future Work}
\label{chap:conclusions}

The work presented in this thesis explores the function of tactile sensing in robotic manipulation. In Chapter \ref{chap:ch3_pose_estimate}, we use the data from the tactile sensors to estimate the object's pose. In Chapter \ref{chap:ch4_pick_objects} we use feedback from tactile sensors to correct for the deviance in object pose estimates obtained from the camera during grasp and in Chapter \ref{chap:ch5_pull_objects} we use a neural network pretrained on human examples and tactile feedback for a Q-learning agent to extract a peg from a hole as a form of disassembly.

We demonstrated that tactile sensors can be beneficial in various aspects of robotic manipulation. They can either support vision based sensors or provide data that vision based sensors can not. The sensors in the phalanges provide localized data at the point of contact with the objects being interfaced. Reinforcement learning is also useful in adapting to dynamic environments and uncertainty. 

In Chapter \ref{chap:ch3_pose_estimate}, we evaluated the effectiveness of a sliding window sampling strategy for pose estimation using tactile data collected from an underactuated gripper equipped with multiple sensors. By employing an LSTM network, we leveraged the temporal nature of the data to improve angle prediction accuracy for objects under grasp manipulation.

The results demonstrated that the use of temporal data from the tactile sensors provided the lowest error. In particular, the window size of 40 samples captured sufficient tactile information to yield the lowest MAE of 0.0375 radians. This performance was consistent across various metrics, including MSE, R$^2$, and EXP. Additionally, our model outperformed traditional regression models, such as ridge and linear regression, which achieved significantly lower R$^2$ and EXP scores, highlighting the advantage of utilizing temporal data for pose estimation.

The successful integration of different sensor readings, synchronized and downsampled to match the camera's frame rate, was crucial for our LSTM model's performance. The preprocessing steps ensured that the diverse sensor data were aligned correctly, facilitating effective model training and evaluation.

This study's findings underscore the potential of using LSTM networks for time-series analysis in tactile sensing applications, particularly in robotics. Future work could explore extending this approach to more complex and varied object shapes and further incorporating additional sensors to enhance pose estimation's robustness and accuracy.

In Chapter \ref{chap:ch4_pick_objects} we develop a system to reduce the the number of grasp attempts based on the uncertainty of the object pose obtained from cameras. We do this by using reinforcement learning to and tactile sensing. 

The tactile sensors obtain local information based on each grasp attempt, which then helps the reinforcement learning agent update its knowledge of the environment. These attempts can help update the policy and get an improved estimate of the uncertainty distribution, which could enable the grasp attempt to be more likely to be successful.  

The total number of grasp attempts tended towards 2 steps from an average of 5 steps after training. This approach enables the manipulator to improve its performance over time, minimizing the number of attempts needed to achieve a successful grasp, even in pose uncertainty. The other benefit of this strategy is the ability to adapt to the changing uncertainty due to environmental factors.

Finally, after obtaining a grasp, we attempt to extract the object in the peg-in-hole problem in \ref{chap:ch5_pull_objects}. We use three different objects, a vertical, a slanted, and a curved peg, to have variability in the trajectory and to determine the transferability and adaptability of the system. The methodology is divided into three main stages: data collection through teleoperation, pre-training a neural network using the teleoperation data, and applying a reinforcement learning (RL) agent with the pre-trained network to perform object extraction.

The pre-training stage uses a dense layer deep neural network to encode user inputs and sensor data, significantly reducing the time required for the RL agent to learn the task. The RL agent employs a deep Q-learning algorithm, optimizing its actions based on a reward function that accounts for pressure changes and end-effector height. Experiments with different peg shapes demonstrate the benefits of pre-training, showing that it reduces the number of episodes needed for successful extraction and improves the generalizability of the learned behaviors across various objects.

Future works should explore joining the in hand pose estimation, grasping and object extraction into one continuous pipeline. It would benefit from shared feature information that can be useful in various combinations of sub-tasks. 

Moreover, increasing the dimensions and degrees of freedom of the manipulator used would be useful in identifying the complexities of scaling. 

Finally, exploring real life applications such as USB connector insertions would be illustrate the feasibility of this work.

\addcontentsline{toc}{chapter}{Bibliography}
\bibliography{ref}

\begin{thebibliography}{100}

\bibitem{Gasparetto2019}
A.~Gasparetto and L.~Scalera.
\newblock A brief history of industrial robotics in the 20th century.
\newblock {\em Advances in historical studies}, 8(1):24--35, 2019.

\bibitem{iavazzo2014evolution}
Christos Iavazzo, Xanthi-Ekaterini~D Gkegke, Paraskevi-Evangelia Iavazzo, and
  Ioannis~D Gkegkes.
\newblock Evolution of robots throughout history from hephaestus to da vinci
  robot.
\newblock {\em Acta medico-historica adriatica: AMHA}, 12(2):247--258, 2014.

\bibitem{zamalloa2017}
Irati Zamalloa, Risto Kojcev, Alejandro Hern{\'a}ndez, Inigo Muguruza, Lander
  Usategui, Asier Bilbao, and Victor Mayoral.
\newblock Dissecting robotics-historical overview and future perspectives.
\newblock {\em arXiv preprint arXiv:1704.08617}, 2017.

\bibitem{kumar2023design}
Ravi Kumar, J{\o}rgen Dale, Jayant Singh, and Jing Zhou.
\newblock Design and development of an anthropomorphic gripper for service
  robotics and prosthetic applications.
\newblock In {\em 2023 11th International Conference on Control, Mechatronics
  and Automation (ICCMA)}, pages 233--238. IEEE, 2023.

\bibitem{Liu2021}
Rongrong Liu, Florent Nageotte, Philippe Zanne, Michel de~Mathelin, and
  Birgitta Dresp-Langley.
\newblock Deep reinforcement learning for the control of robotic manipulation:
  A focussed mini-review.
\newblock {\em Robotics}, 10(1), 2021.

\bibitem{dou2021soft}
Weiqiang Dou, Guoliang Zhong, Jinglin Cao, Zhun Shi, Bowen Peng, and Liangzhong
  Jiang.
\newblock Soft robotic manipulators: Designs, actuation, stiffness tuning, and
  sensing.
\newblock {\em Advanced Materials Technologies}, 6(9):2100018, 2021.

\bibitem{helbig2008haptic}
Hannah~B Helbig and Marc~O Ernst.
\newblock Haptic perception in interaction with other senses.
\newblock In {\em Human haptic perception: Basics and applications}, pages
  235--249. Springer, 2008.

\bibitem{Li2020}
Shuang Li, Jiaxi Jiang, Philipp Ruppel, Hongzhuo Liang, Xiaojian Ma, Norman
  Hendrich, Fuchun Sun, and Jianwei Zhang.
\newblock A mobile robot hand-arm teleoperation system by vision and imu.
\newblock In {\em 2020 IEEE/RSJ International Conference on Intelligent Robots
  and Systems (IROS)}, pages 10900--10906, 2020.

\bibitem{Alatise2020}
Mary~B. Alatise and Gerhard~P. Hancke.
\newblock A review on challenges of autonomous mobile robot and sensor fusion
  methods.
\newblock {\em IEEE Access}, 8:39830--39846, 2020.

\bibitem{wang2021gelsight}
Shaoxiong Wang, Yu~She, Branden Romero, and Edward Adelson.
\newblock Gelsight wedge: Measuring high-resolution 3d contact geometry with a
  compact robot finger.
\newblock In {\em 2021 IEEE International Conference on Robotics and Automation
  (ICRA)}, pages 6468--6475. IEEE, 2021.

\bibitem{dong2021tactile}
Siyuan Dong, Devesh~K Jha, Diego Romeres, Sangwoon Kim, Daniel Nikovski, and
  Alberto Rodriguez.
\newblock Tactile-rl for insertion: Generalization to objects of unknown
  geometry.
\newblock In {\em 2021 IEEE International Conference on Robotics and Automation
  (ICRA)}, pages 6437--6443. IEEE, 2021.

\bibitem{DuGuoguang2021Vrgf}
Guoguang Du, Kai Wang, Shiguo Lian, and Kaiyong Zhao.
\newblock Vision-based robotic grasping from object localization, object pose
  estimation to grasp estimation for parallel grippers: a review.
\newblock {\em The Artificial intelligence review}, 54(3):1677--1734, 2021.

\bibitem{Sipos2022}
Andrea Sipos and Nima Fazeli.
\newblock Simultaneous contact location and object pose estimation using
  proprioception and tactile feedback.
\newblock In {\em 2022 IEEE/RSJ International Conference on Intelligent Robots
  and Systems (IROS)}, pages 3233--3240, 2022.

\bibitem{Li2021}
Yiming Li, Tao Kong, Ruihang Chu, Yifeng Li, Peng Wang, and Lei Li.
\newblock Simultaneous semantic and collision learning for 6-dof grasp pose
  estimation.
\newblock In {\em 2021 IEEE/RSJ International Conference on Intelligent Robots
  and Systems (IROS)}, pages 3571--3578, 2021.

\bibitem{contact_pose}
Felix von Drigalski, Shohei Taniguchi, Robert Lee, Takamitsu Matsubara, Masashi
  Hamaya, Kazutoshi Tanaka, and Yoshihisa Ijiri.
\newblock Contact-based in-hand pose estimation using bayesian state estimation
  and particle filtering.
\newblock 05 2020.

\bibitem{bauza2023tac2pose}
Maria Bauza, Antonia Bronars, and Alberto Rodriguez.
\newblock Tac2pose: Tactile object pose estimation from the first touch.
\newblock {\em The International Journal of Robotics Research},
  42(13):1185--1209, 2023.

\bibitem{Yang2024}
Sanghoon Yang, Won~Dong Kim, Hyunkyu Park, Seojung Min, Hyonyoung Han, and Jung
  Kim.
\newblock In-hand object classification and pose estimation with sim-to-real
  tactile transfer for robotic manipulation.
\newblock {\em IEEE Robotics and Automation Letters}, 9(1):659--666, 2024.

\bibitem{kelestemur2022tactile}
Tarik Kelestemur, Robert Platt, and Taskin Padir.
\newblock Tactile pose estimation and policy learning for unknown object
  manipulation.
\newblock In {\em Proceedings of the 21st International Conference on
  Autonomous Agents and Multiagent Systems}, pages 742--750, 2022.

\bibitem{Jentoft2014}
Leif~P. Jentoft, Qian Wan, and Robert~D. Howe.
\newblock Limits to compliance and the role of tactile sensing in grasping.
\newblock In {\em 2014 IEEE International Conference on Robotics and Automation
  (ICRA)}, pages 6394--6399, 2014.

\bibitem{Xiong2022}
Wennan Xiong, Hui Feng, Haosen Liwang, Dan Li, Wanbing Yao, Dilinazha Duolikun,
  Yunlei Zhou, and YongAn Huang.
\newblock Multifunctional tactile feedbacks towards compliant robot
  manipulations via 3d-shaped electronic skin.
\newblock {\em IEEE Sensors Journal}, 22(9):9046--9056, 2022.

\bibitem{Fernandez2021}
Alfonso~J. Fernandez, Huan Weng, Paul~B. Umbanhowar, and Kevin~M. Lynch.
\newblock Visiflex: A low-cost compliant tactile fingertip for force, torque,
  and contact sensing.
\newblock {\em IEEE Robotics and Automation Letters}, 6(2):3009--3016, 2021.

\bibitem{Liu2022}
Sandra~Q. Liu and Edward~H. Adelson.
\newblock Gelsight fin ray: Incorporating tactile sensing into a soft compliant
  robotic gripper.
\newblock In {\em 2022 IEEE 5th International Conference on Soft Robotics
  (RoboSoft)}, pages 925--931, 2022.

\bibitem{Oliveira2017}
Thiago~Eustaquio Alves~de Oliveira, Ana-Maria Cretu, and Emil~M. Petriu.
\newblock Multimodal bio-inspired tactile sensing module.
\newblock {\em IEEE Sensors Journal}, 17(11):3231--3243, 2017.

\bibitem{cretu2015computational}
Ana-Maria Cretu, Thiago Eustaquio~Alves De~Oliveira, Vinicius~Prado Da~Fonseca,
  Bilal Tawbe, Emil~M Petriu, and Voicu~Z Groza.
\newblock Computational intelligence and mechatronics solutions for robotic
  tactile object recognition.
\newblock In {\em 2015 IEEE 9th international symposium on intelligent signal
  processing (WISP) proceedings}, pages 1--6. IEEE, 2015.

\bibitem{de2023bioin}
Thiago Eustaquio~Alves de~Oliveira and Vinicius~Prado da~Fonseca.
\newblock Bioin-tacto: A compliant multi-modal tactile sensing module for
  robotic tasks.
\newblock {\em HardwareX}, 16:e00478, 2023.

\bibitem{Kuppuswamy2020}
Naveen Kuppuswamy, Alejandro Castro, Calder Phillips-Grafflin, Alex Alspach,
  and Russ Tedrake.
\newblock Fast model-based contact patch and pose estimation for highly
  deformable dense-geometry tactile sensors.
\newblock {\em IEEE Robotics and Automation Letters}, 5:1811--1818, 4 2020.

\bibitem{Jiang2023}
Jiaqi Jiang, Guanqun Cao, Aaron Butterworth, Thanh-Toan Do, and Shan Luo.
\newblock Where shall i touch? vision-guided tactile poking for transparent
  object grasping.
\newblock {\em IEEE/ASME Transactions on Mechatronics}, 28(1):233--244, 2023.

\bibitem{jiang2020state}
Jingang Jiang, Zhiyuan Huang, Zhuming Bi, Xuefeng Ma, and Guang Yu.
\newblock State-of-the-art control strategies for robotic pih assembly.
\newblock {\em Robotics and Computer-Integrated Manufacturing}, 65:101894,
  2020.

\bibitem{dong2019tactile}
Siyuan Dong and Alberto Rodriguez.
\newblock Tactile-based insertion for dense box-packing.
\newblock In {\em 2019 IEEE/RSJ International Conference on Intelligent Robots
  and Systems (IROS)}, pages 7953--7960. IEEE, 2019.

\bibitem{Shiyu2021}
Shiyu Jin, Xinghao Zhu, Changhao Wang, and Masayoshi Tomizuka.
\newblock Contact pose identification for peg-in-hole assembly under
  uncertainties.
\newblock In {\em 2021 American Control Conference (ACC)}, pages 48--53, 2021.

\bibitem{SONG20211}
Jingzhou Song, Qingle Chen, and Zhendong Li.
\newblock A peg-in-hole robot assembly system based on gauss mixture model.
\newblock {\em Robotics and Computer-Integrated Manufacturing}, 67:101996,
  2021.

\bibitem{Lee2022}
Dong-Hyuk Lee, Myoung-Su Choi, Hyeonjun Park, Ga-Ram Jang, Jae-Han Park, and
  Ji-Hun Bae.
\newblock Peg-in-hole assembly with dual-arm robot and dexterous robot hands.
\newblock {\em IEEE Robotics and Automation Letters}, 7(4):8566--8573, 2022.

\bibitem{Kang2022}
Hanwen Kang, Yaohua Zang, Xing Wang, and Yaohui Chen.
\newblock Uncertainty-driven spiral trajectory for robotic peg-in-hole
  assembly.
\newblock {\em IEEE Robotics and Automation Letters}, 7(3):6661--6668, 2022.

\bibitem{kim2022}
Sangwoon Kim and Alberto Rodriguez.
\newblock Active extrinsic contact sensing: Application to general peg-in-hole
  insertion.
\newblock In {\em 2022 International Conference on Robotics and Automation
  (ICRA)}, pages 10241--10247, 2022.

\bibitem{Galaiya2023}
Viral~Rasik Galaiya, Mohammed Asfour, Thiago~Eustaquio Alves~de Oliveira,
  Xianta Jiang, and Vinicius Prado~da Fonseca.
\newblock Exploring tactile temporal features for object pose estimation during
  robotic manipulation.
\newblock {\em Sensors}, 23(9):4535, May 2023.

\bibitem{billard2019}
Baude Billard and Danica Kragic.
\newblock Trends and challenges in robot manipulation.
\newblock {\em Science}, 364(6446):eaat8414, 2019.

\bibitem{Hammond2012}
Frank~L. Hammond, Jonathan Weisz, Andrés~A. de~la Llera~Kurth, Peter~K. Allen,
  and Robert~D. Howe.
\newblock Towards a design optimization method for reducing the mechanical
  complexity of underactuated robotic hands.
\newblock In {\em 2012 IEEE International Conference on Robotics and
  Automation}, pages 2843--2850, 2012.

\bibitem{SAHIN2020103898}
Caner Sahin, Guillermo Garcia-Hernando, Juil Sock, and Tae-Kyun Kim.
\newblock A review on object pose recovery: From 3d bounding box detectors to
  full 6d pose estimators.
\newblock {\em Image and Vision Computing}, 96:103898, 2020.

\bibitem{Kroemer2019}
Julian Zimmer, Tess Hellebrekers, Tamim Asfour, Carmel Majidi, and Oliver
  Kroemer.
\newblock Predicting grasp success with a soft sensing skin and shape-memory
  actuated gripper.
\newblock In {\em 2019 IEEE/RSJ International Conference on Intelligent Robots
  and Systems (IROS)}, pages 7120--7127, 2019.

\bibitem{wettels2008biomimetic}
Nicholas Wettels, Veronica~J Santos, Roland~S Johansson, and Gerald~E Loeb.
\newblock Biomimetic tactile sensor array.
\newblock {\em Advanced robotics}, 22(8):829--849, 2008.

\bibitem{ward2018tactip}
Benjamin Ward-Cherrier, Nicholas Pestell, Luke Cramphorn, Benjamin Winstone,
  Maria~Elena Giannaccini, Jonathan Rossiter, and Nathan~F Lepora.
\newblock The tactip family: Soft optical tactile sensors with 3d-printed
  biomimetic morphologies.
\newblock {\em Soft robotics}, 5(2):216--227, 2018.

\bibitem{lambeta2020digit}
Mike Lambeta, Po-Wei Chou, Stephen Tian, Brian Yang, Benjamin Maloon,
  Victoria~Rose Most, Dave Stroud, Raymond Santos, Ahmad Byagowi, Gregg
  Kammerer, et~al.
\newblock Digit: A novel design for a low-cost compact high-resolution tactile
  sensor with application to in-hand manipulation.
\newblock {\em IEEE Robotics and Automation Letters}, 5(3):3838--3845, 2020.

\bibitem{alspach2019soft}
Alex Alspach, Kunimatsu Hashimoto, Naveen Kuppuswamy, and Russ Tedrake.
\newblock Soft-bubble: A highly compliant dense geometry tactile sensor for
  robot manipulation.
\newblock In {\em 2019 2nd IEEE International Conference on Soft Robotics
  (RoboSoft)}, pages 597--604. IEEE, 2019.

\bibitem{su2015force}
Zhe Su, Karol Hausman, Yevgen Chebotar, Artem Molchanov, Gerald~E Loeb,
  Gaurav~S Sukhatme, and Stefan Schaal.
\newblock Force estimation and slip detection/classification for grip control
  using a biomimetic tactile sensor.
\newblock In {\em 2015 IEEE-RAS 15th International Conference on Humanoid
  Robots (Humanoids)}, pages 297--303. IEEE, 2015.

\bibitem{yoon2022Elong}
Sohee~John Yoon, Minsik Choi, Bomin Jeong, and Yong-Lae Park.
\newblock Elongatable gripper fingers with integrated stretchable tactile
  sensors for underactuated grasping and dexterous manipulation.
\newblock {\em IEEE Transactions on Robotics}, 38(4):2179--2193, 2022.

\bibitem{Bandari2020}
Naghmeh Bandari, Javad Dargahi, and Muthukumaran Packirisamy.
\newblock Tactile sensors for minimally invasive surgery: A review of the
  state-of-the-art, applications, and perspectives.
\newblock {\em IEEE Access}, 8:7682--7708, 2020.

\bibitem{she2021cable}
Yu~She, Shaoxiong Wang, Siyuan Dong, Neha Sunil, Alberto Rodriguez, and Edward
  Adelson.
\newblock Cable manipulation with a tactile-reactive gripper.
\newblock {\em The International Journal of Robotics Research},
  40(12-14):1385--1401, 2021.

\bibitem{bi2021zero}
Thomas Bi, Carmelo Sferrazza, and Raffaello D’Andrea.
\newblock Zero-shot sim-to-real transfer of tactile control policies for
  aggressive swing-up manipulation.
\newblock {\em IEEE Robotics and Automation Letters}, 6(3):5761--5768, 2021.

\bibitem{zhang2023interaction}
Hanwen Zhang, Zeyu Lu, Wenyu Liang, Haoyong Yu, Yao Mao, and Yan Wu.
\newblock Interaction control for tool manipulation on deformable objects using
  tactile feedback.
\newblock {\em IEEE Robotics and Automation Letters}, 2023.

\bibitem{VonDrigalski2020}
Felix~Von Drigalski, Shohei Taniguchi, Robert Lee, Takamitsu Matsubara, Masashi
  Hamaya, Kazutoshi Tanaka, and Yoshihisa Ijiri.
\newblock Contact-based in-hand pose estimation using bayesian state estimation
  and particle filtering.
\newblock {\em Proceedings - IEEE International Conference on Robotics and
  Automation}, pages 7294--7299, 2020.

\bibitem{Gomes2020}
Daniel~Fernandes Gomes, Zhonglin Lin, and Shan Luo.
\newblock Geltip: A finger-shaped optical tactile sensor for robotic
  manipulation.
\newblock In {\em 2020 IEEE/RSJ International Conference on Intelligent Robots
  and Systems (IROS)}, pages 9903--9909, 2020.

\bibitem{Romero2020}
Branden Romero, Filipe Veiga, and Edward Adelson.
\newblock Soft, round, high resolution tactile fingertip sensors for dexterous
  robotic manipulation.
\newblock In {\em 2020 IEEE International Conference on Robotics and Automation
  (ICRA)}, pages 4796--4802, 2020.

\bibitem{Trueeb2020}
Camill Trueeb, Carmelo Sferrazza, and Raffaello D’Andrea.
\newblock Towards vision-based robotic skins: a data-driven, multi-camera
  tactile sensor.
\newblock In {\em 2020 3rd IEEE International Conference on Soft Robotics
  (RoboSoft)}, pages 333--338, 2020.

\bibitem{PradodaFonseca2019}
Vinicius~Prado da~Fonseca, Thiago Eustaquio~Alves de~Oliveira, and Emil~M.
  Petriu.
\newblock the orientation of objects from tactile sensing data using machine
  learning methods and visual frames of reference.
\newblock {\em Sensors (Basel, Switzerland)}, 19, 5 2019.

\bibitem{Sipos2022sim}
Andrea Sipos and Nima Fazeli.
\newblock Simultaneous contact location and object pose estimation using
  proprioception and tactile feedback.
\newblock In {\em 2022 IEEE/RSJ International Conference on Intelligent Robots
  and Systems (IROS)}, pages 3233--3240, 2022.

\bibitem{Lloyd2022goal}
John Lloyd and Nathan~F. Lepora.
\newblock Goal-driven robotic pushing using tactile and proprioceptive
  feedback.
\newblock {\em IEEE Transactions on Robotics}, 38(2):1201--1212, 2022.

\bibitem{Lima2019zscore}
Bruno Monteiro~Rocha Lima, Luiz Claudio~Sampaio Ramos, Thiago Eustaquio~Alves
  de~Oliveira, Vinicius~Prado da~Fonseca, and Emil~M. Petriu.
\newblock Heart rate detection using a multimodal tactile sensor and a z-score
  based peak detection algorithm.
\newblock {\em CMBES Proceedings}, 42, May 2019.

\bibitem{de2015data}
Thiago Eustaquio~Alves de~Oliveira, Vinicius~Prado da~Fonseca, Emanuil Huluta,
  Paulo~FF Rosa, and Emil~M Petriu.
\newblock Data-driven analysis of kinaesthetic and tactile information for
  shape classification.
\newblock In {\em 2015 IEEE International Conference on Computational
  Intelligence and Virtual Environments for Measurement Systems and
  Applications (CIVEMSA)}, pages 1--5. IEEE, 2015.

\bibitem{new_ji_tactile}
Jingjing Ji, Yuting Liu, and Huan Ma.
\newblock Model-based 3d contact geometry perception for visual tactile sensor.
\newblock {\em Sensors}, 22(17), 2022.

\bibitem{new_vts}
Umer Shah, Rajkumar Muthusamy, Dongming Gan, Yahya Zweiri, and Lakmal
  Seneviratne.
\newblock On the design and development of vision-based tactile sensors.
\newblock {\em Journal of Intelligent \& Robotic Systems}, 102, 08 2021.

\bibitem{new_suresh_slam}
Sudharshan Suresh, Maria Bauza, Kuan-Ting Yu, Joshua~G. Mangelson, Alberto
  Rodriguez, and Michael Kaess.
\newblock Tactile slam: Real-time inference of shape and pose from planar
  pushing.
\newblock In {\em 2021 IEEE International Conference on Robotics and Automation
  (ICRA)}, page 11322–11328. IEEE Press, 2021.

\bibitem{alvarez}
David {\'A}lvarez, M{\'a}ximo~A. Roa, and Luis Moreno.
\newblock Tactile-based in-hand object pose estimation.
\newblock In Anibal Ollero, Alberto Sanfeliu, Luis Montano, Nuno Lau, and
  Carlos Cardeira, editors, {\em ROBOT 2017: Third Iberian Robotics
  Conference}, pages 716--728, Cham, 2018. Springer International Publishing.

\bibitem{Azulay}
Osher Azulay, Inbar Ben-David, and Avishai Sintov.
\newblock Learning haptic-based object pose estimation for in-hand manipulation
  with underactuated robotic hands.
\newblock {\em arXiv}, 2022.

\bibitem{Funaabashi2022multi}
Satoshi Funabashi, Tomoki Isobe, Fei Hongyi, Atsumu Hiramoto, Alexander
  Schmitz, Shigeki Sugano, and Tetsuya Ogata.
\newblock Multi-fingered in-hand manipulation with various object properties
  using graph convolutional networks and distributed tactile sensors.
\newblock {\em IEEE Robotics and Automation Letters}, 7(2):2102--2109, 2022.

\bibitem{new_alvarez_fusion}
David {\'A}lvarez, M{\'a}ximo~A. Roa, and Luis Moreno.
\newblock Visual and tactile fusion for estimating the pose of a grasped
  object.
\newblock In Manuel~F. Silva, Jos{\'e} Lu{\'i}s~Lima, Lu{\'i}s~Paulo Reis,
  Alberto Sanfeliu, and Danilo Tardioli, editors, {\em Robot 2019: Fourth
  Iberian Robotics Conference}, pages 184--198, Cham, 2020. Springer
  International Publishing.

\bibitem{new_dikhale_inhand}
Snehal Dikhale, Karankumar Patel, Daksh Dhingra, Itoshi Naramura, Akinobu
  Hayashi, Soshi Iba, and Nawid Jamali.
\newblock Visuotactile 6d pose estimation of an in-hand object using vision and
  tactile sensor data.
\newblock {\em IEEE Robotics and Automation Letters}, 7(2):2148--2155, 2022.

\bibitem{ros}
{Stanford Artificial Intelligence Laboratory et al.}
\newblock Robotic operating system.

\bibitem{PradodaFonseca2022tactile}
Vinicius~Prado da~Fonseca, Xianta Jiang, Emil~M Petriu, and Thiago
  Eustaquio~Alves de~Oliveira.
\newblock Tactile object recognition in early phases of grasping using
  underactuated robotic hands.
\newblock {\em Intelligent Service Robotics}, 15(4):513--525, 2022.

\bibitem{nigatu2021analysis}
Hassen Nigatu, Yun~Ho Choi, and Doik Kim.
\newblock Analysis of parasitic motion with the constraint embedded jacobian
  for a 3-prs parallel manipulator.
\newblock {\em Mechanism and Machine Theory}, 164:104409, 2021.

\bibitem{tensorflow2015-whitepaper}
Mart\'{i}n Abadi, Ashish Agarwal, Paul Barham, Eugene Brevdo, Zhifeng Chen,
  Craig Citro, Greg~S. Corrado, Andy Davis, Jeffrey Dean, Matthieu Devin,
  Sanjay Ghemawat, Ian Goodfellow, Andrew Harp, Geoffrey Irving, Michael Isard,
  Yangqing Jia, Rafal Jozefowicz, Lukasz Kaiser, Manjunath Kudlur, Josh
  Levenberg, Dandelion Man\'{e}, Rajat Monga, Sherry Moore, Derek Murray, Chris
  Olah, Mike Schuster, Jonathon Shlens, Benoit Steiner, Ilya Sutskever, Kunal
  Talwar, Paul Tucker, Vincent Vanhoucke, Vijay Vasudevan, Fernanda Vi\'{e}gas,
  Oriol Vinyals, Pete Warden, Martin Wattenberg, Martin Wicke, Yuan Yu, and
  Xiaoqiang Zheng.
\newblock {TensorFlow}: Large-scale machine learning on heterogeneous systems,
  2015.
\newblock Software available from tensorflow.org.

\bibitem{Mathew2018}
Matthew~T. Mason.
\newblock Toward robotic manipulation.
\newblock {\em Annual Review of Control, Robotics, and Autonomous Systems},
  1(1):1--28, 2018.

\bibitem{Hao2023}
Wenhan Hao, Chen Zhu, and Michael Meurer.
\newblock Camera calibration error modeling and its impact on visual
  positioning.
\newblock In {\em 2023 IEEE/ION Position, Location and Navigation Symposium
  (PLANS)}, pages 1394--1399, 2023.

\bibitem{Hoda2016}
Tom{\'a}{\v{s}} Hoda{\v{n}}, Ji{\v{r}}{\'i} Matas, and {\v{S}}t{\v{e}}p{\'a}n
  Obdr{\v{z}}{\'a}lek.
\newblock On evaluation of 6d object pose estimation.
\newblock In Gang Hua and Herv{\'e} J{\'e}gou, editors, {\em Computer Vision --
  ECCV 2016 Workshops}, pages 606--619, Cham, 2016. Springer International
  Publishing.

\bibitem{jiang2022shall}
Jiaqi Jiang, Guanqun Cao, Aaron Butterworth, Thanh-Toan Do, and Shan Luo.
\newblock Where shall i touch? vision-guided tactile poking for transparent
  object grasping.
\newblock {\em IEEE/ASME Transactions on Mechatronics}, 28(1):233--244, 2022.

\bibitem{prado2019estimating}
Vinicius Prado~da Fonseca, Thiago~Eustaquio Alves~de Oliveira, and Emil~M
  Petriu.
\newblock Estimating the orientation of objects from tactile sensing data using
  machine learning methods and visual frames of reference.
\newblock {\em Sensors}, 19(10):2285, 2019.

\bibitem{Lima2021}
Bruno~Monteiro Rocha~Lima, Thiago~Eustaquio Alves~de Oliveira, and
  Vinicius~Prado da~Fonseca.
\newblock Classification of textures using a tactile-enabled finger in dynamic
  exploration tasks.
\newblock In {\em 2021 IEEE Sensors}, pages 1--4, 2021.

\bibitem{lima2020dynamic}
Bruno Monteiro~Rocha Lima, Vinicius~Prado da~Fonseca, Thiago Eustaquio~Alves
  de~Oliveira, Qi~Zhu, and Emil~M Petriu.
\newblock Dynamic tactile exploration for texture classification using a
  miniaturized multi-modal tactile sensor and machine learning.
\newblock In {\em 2020 IEEE International Systems Conference (SysCon)}, pages
  1--7. IEEE, 2020.

\bibitem{Pohtongkam2023}
Somchai Pohtongkam and Jakkree Srinonchat.
\newblock Object recognition for humanoid robots using full hand tactile
  sensor.
\newblock {\em IEEE Access}, 11:20284--20297, 2023.

\bibitem{da2022tactile}
Vinicius~Prado da~Fonseca, Xianta Jiang, Emil~M Petriu, and Thiago
  Eustaquio~Alves de~Oliveira.
\newblock Tactile object recognition in early phases of grasping using
  underactuated robotic hands.
\newblock {\em Intelligent Service Robotics}, 15(4):513--525, 2022.

\bibitem{Mandil2023}
Willow Mandil, Vishnu Rajendran, Kiyanoush Nazari, and Amir Ghalamzan-Esfahani.
\newblock Tactile-sensing technologies: Trends, challenges and outlook in
  agri-food manipulation.
\newblock {\em Sensors}, 23(17), 2023.

\bibitem{wang2020feature}
Chao Wang, Xuehe Zhang, Xizhe Zang, Yubin Liu, Guanwen Ding, Wenxin Yin, and
  Jie Zhao.
\newblock Feature sensing and robotic grasping of objects with uncertain
  information: A review.
\newblock {\em Sensors}, 20(13):3707, 2020.

\bibitem{Kleeberger2020}
Kilian Kleeberger, Richard Bormann, Werner Kraus, and Marco Huber.
\newblock A survey on learning-based robotic grasping.
\newblock {\em Current Robotics Reports}, 1:239–249, 12 2020.

\bibitem{James2017}
Stephen James, Andrew Davison, and Edward Johns.
\newblock Transferring end-to-end visuomotor control from simulation to real
  world for a multi-stage task.
\newblock {\em Conference on Robot Learning}, 07 2017.

\bibitem{matak2022planning}
Martin Matak and Tucker Hermans.
\newblock Planning visual-tactile precision grasps via complementary use of
  vision and touch.
\newblock {\em IEEE Robotics and Automation Letters}, 8(2):768--775, 2022.

\bibitem{gelslim}
I.~{Taylor}, S.~{Dong}, and A.~{Rodriguez}.
\newblock Gelslim3.0: High-resolution measurement of shape, force and slip in a
  compact tactile-sensing finger, 2021.

\bibitem{welyhorsky2022neuro}
Maxwell Welyhorsky, Vinicius~Prado Da~Fonseca, Qi~Zhu, Bruno Monteiro~Rocha
  Lima, Thiago Eustaquio~Alves De~Oliveira, and Emil~M Petriu.
\newblock Neuro-fuzzy grasp control for a teleoperated five finger
  anthropomorphic robotic hand.
\newblock In {\em 2022 IEEE International Systems Conference (SysCon)}, pages
  1--5. IEEE, 2022.

\bibitem{zhu2020teleoperated}
Qi~Zhu, Vinicius~Prado da~Fonseca, Bruno Monteiro~Rocha Lima, Maxwell
  Welyhorsky, Miriam Goubran, Thiago Eustaquio~Alves de~Oliveira, and Emil~M
  Petriu.
\newblock Teleoperated grasping using a robotic hand and a haptic-feedback data
  glove.
\newblock In {\em 2020 IEEE International Systems Conference (SysCon)}, pages
  1--7. IEEE, 2020.

\bibitem{beltran2020variable}
Cristian~C Beltran-Hernandez, Damien Petit, Ixchel~G Ramirez-Alpizar, and
  Kensuke Harada.
\newblock Variable compliance control for robotic peg-in-hole assembly: A
  deep-reinforcement-learning approach.
\newblock {\em Applied Sciences}, 10(19):6923, 2020.

\bibitem{unten2022peg}
Hikaru Unten, Sho Sakaino, and Toshiaki Tsuji.
\newblock Peg-in-hole using transient information of force response.
\newblock {\em IEEE/ASME Transactions on Mechatronics}, 2022.

\bibitem{nigro2020peg}
Michelangelo Nigro, Monica Sileo, Francesco Pierri, Katia Genovese, Domenico~D
  Bloisi, and Fabrizio Caccavale.
\newblock Peg-in-hole using 3d workpiece reconstruction and cnn-based hole
  detection.
\newblock In {\em 2020 IEEE/RSJ International Conference on Intelligent Robots
  and Systems (IROS)}, pages 4235--4240. IEEE, 2020.

\bibitem{Jin2021}
Shiyu Jin, Xinghao Zhu, Changhao Wang, and Masayoshi Tomizuka.
\newblock Contact pose identification for peg-in-hole assembly under
  uncertainties.
\newblock In {\em 2021 American Control Conference (ACC)}, pages 48--53, 2021.

\bibitem{SONG2021101996}
Jingzhou Song, Qingle Chen, and Zhendong Li.
\newblock A peg-in-hole robot assembly system based on gauss mixture model.
\newblock {\em Robotics and Computer-Integrated Manufacturing}, 67:101996,
  2021.

\bibitem{wang2021alignment}
Yongzhi Wang, Lei Zhao, Qian Zhang, Ran Zhou, Liping Wu, Junqiao Ma, Bo~Zhang,
  and Yu~Zhang.
\newblock Alignment method of combined perception for peg-in-hole assembly with
  deep reinforcement learning.
\newblock {\em Journal of Sensors}, 2021:1--12, 2021.

\bibitem{LU2022}
Yuqian Lu, Hao Zheng, Saahil Chand, Wanqing Xia, Zengkun Liu, Xun Xu, Lihui
  Wang, Zhaojun Qin, and Jinsong Bao.
\newblock Outlook on human-centric manufacturing towards industry 5.0.
\newblock {\em Journal of Manufacturing Systems}, 62:612--627, 2022.

\bibitem{LEE2024}
Regina Kyung-Jin Lee, Hao Zheng, and Yuqian Lu.
\newblock Human-robot shared assembly taxonomy: A step toward seamless
  human-robot knowledge transfer.
\newblock {\em Robotics and Computer-Integrated Manufacturing}, 86:102686,
  2024.

\bibitem{lammle2022}
Arik Lämmle, Philipp Tenbrock, Balázs Bálint, Frank Nägele, Werner Kraus,
  József Váncza, and Marco~F. Huber.
\newblock Simulation-based learning of the peg-in-hole process using
  robot-skills.
\newblock In {\em 2022 IEEE/RSJ International Conference on Intelligent Robots
  and Systems (IROS)}, pages 9340--9346, 2022.

\bibitem{Kozlovsky2022}
Shir Kozlovsky, Elad Newman, and Miriam Zacksenhouse.
\newblock Reinforcement learning of impedance policies for peg-in-hole tasks:
  Role of asymmetric matrices.
\newblock {\em IEEE Robotics and Automation Letters}, 7(4):10898--10905, 2022.

\bibitem{Beltran2020}
Cristian~C. Beltran-Hernandez, Damien Petit, Ixchel~G. Ramirez-Alpizar, and
  Kensuke Harada.
\newblock Variable compliance control for robotic peg-in-hole assembly: A
  deep-reinforcement-learning approach.
\newblock {\em Applied Sciences}, 10(19), 2020.

\bibitem{de2019end}
Thiago Eustaquio~Alves de~Oliveira, Vinicius~Prado da~Fonseca, Bruno
  Monteiro~Rocha Lima, Ana-Maria Cretu, and M~Petriu.
\newblock End-effector approach flexibilization in a surface approximation task
  using a bioinspired tactile sensing module.
\newblock In {\em 2019 IEEE International Symposium on Robotic and Sensors
  Environments (ROSE)}, pages 1--6. IEEE, 2019.

\bibitem{xie2023learning}
Liang Xie, Hongxiang Yu, Kechun Xu, Tong Yang, Minhang Wang, Haojian Lu, Rong
  Xiong, and Yue Wang.
\newblock Learning a simulation-based visual policy for real-world peg in
  unseen holes.
\newblock {\em Review of Scientific Instruments}, 94(10), 2023.

\bibitem{mnih2013playing}
Volodymyr Mnih, Koray Kavukcuoglu, David Silver, Alex Graves, Ioannis
  Antonoglou, Daan Wierstra, and Martin Riedmiller.
\newblock Playing atari with deep reinforcement learning.
\newblock {\em arXiv preprint arXiv:1312.5602}, 2013.

\end{thebibliography}


\end{document}